\pdfoutput=1

\documentclass[11pt]{article}

\usepackage[final]{arxiv}

\usepackage{times}
\usepackage{latexsym}
\usepackage[T1]{fontenc}
\usepackage{graphicx}
\usepackage{caption}
\usepackage{subcaption}
\usepackage[linesnumbered,ruled,vlined]{algorithm2e}
\SetKwComment{Comment}{/* }{ */}
\usepackage{enumitem}
\SetKwComment{LeftComment}{\hfill\makebox[5pt][l]{/* } }{ */}

\newcommand\figref[1]{Figure ~\ref{#1}}
\newcommand\secref[1]{Section ~\ref{#1}}
\newcommand\tabref[1]{Table ~\ref{#1}}
\usepackage{multirow}
\usepackage{hyperref}
\usepackage{url}
\usepackage{soul}   
\usepackage[textwidth=2.5cm]{todonotes}
\usepackage{amsmath, amsthm, amssymb}
\usepackage{wrapfig}
\usepackage{nicefrac}
\usepackage{enumitem}
\usepackage{float} %
\usepackage{booktabs}
\usepackage{amssymb}%
\usepackage{pifont}%
\newcommand{\cmark}{\ding{51}}%
\newcommand{\xmark}{\ding{55}}%

\usepackage[utf8]{inputenc}

\usepackage{microtype}

\usepackage{inconsolata}
\usepackage{arydshln}
\usepackage{booktabs}
\usepackage[flushleft]{threeparttable}

\title{ LLM \emph{in a flash}:\\ Efficient Large Language Model Inference with Limited Memory} 
\author{Keivan Alizadeh, Iman Mirzadeh\thanks{\, \,Major Contribution} , Dmitry Belenko\footnotemark[1]  ,  {\bf S. Karen Khatamifard}, \\ {\bf Minsik Cho,}    {\bf Carlo C Del Mundo,}   {\bf Mohammad Rastegari,} {\bf Mehrdad Farajtabar} \\
Apple \thanks{\, {\{kalizadehvahid, imirzadeh, d\_belenko, skhatamifard, minsik, cdelmundo, mrastegari, farajtabar\}@apple.com}}\, }

\begin{document}
\maketitle
\begin{abstract}
Large language models (LLMs) are central to modern natural language processing, delivering exceptional performance in various tasks. However, their substantial computational and memory requirements present challenges, especially for devices with limited DRAM capacity. This paper tackles the challenge of efficiently running LLMs that exceed the available DRAM capacity by storing the model parameters in flash memory, but bringing them on demand to DRAM. 
Our method involves constructing an inference cost model that takes into account the characteristics of flash memory, guiding us to optimize in two critical areas: reducing the volume of data transferred from flash and reading data in larger, more contiguous chunks. Within this hardware-informed framework, we introduce two principal techniques. First, ``windowing'' strategically reduces data transfer by reusing previously activated neurons, and second, ``row-column bundling'', tailored to the sequential data access strengths of flash memory, increases the size of data chunks read from flash memory. These methods collectively enable running models up to twice the size of the available DRAM, with up to 4x and 20x increase in inference speed compared to naive loading approaches in CPU and GPU, respectively. Our integration of sparsity awareness, context-adaptive loading, and a hardware-oriented design paves the way for effective inference of LLMs on devices with limited memory.

\end{abstract}

\section{Introduction}
\label{sec:intro}
In recent years, large language models (LLMs) have demonstrated strong performance across a wide range of natural language tasks~\citep{brown2020language,chowdhery2022palm,Llamav1paper,mistral_paper,team2023gemini}. However, the unprecedented capabilities of these models come with substantial computational and memory requirements for inference. LLMs can contain hundreds of billions or even trillions of parameters, which makes them challenging to load and run efficiently, especially on personal devices.

Currently, the standard approach is to load the entire model into DRAM (Dynamic Random Access Memory) for inference \citep{rajbhandari2021zero, aminabadi2022deepspeed}. However, this severely limits the maximum model size that can be run. For example, a 7 billion parameter model requires over 14GB of memory just to load the parameters in half-precision floating point format, exceeding the capabilities of most personal devices such as smartphones. While it is possible to employ techniques such as quantization to reduce the model size, still, this cannot address the main limitation of loading the entire model into DRAM.

\begin{figure}[t]
\centering
\includegraphics[width=0.99\linewidth]{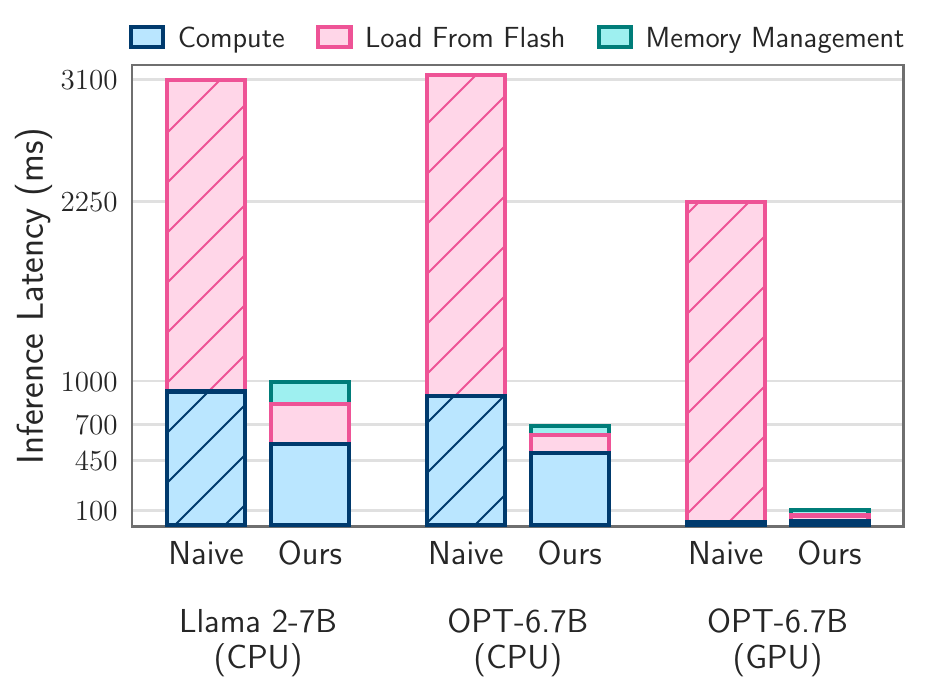}
\caption{Average inference latency for a single token when only half of the model's memory is available: Our method selectively loads parameters on demand for each token generation step. The latency represents the time required to repeatedly load parameters from flash memory, combined with the time needed for computations.}
\label{fig:latency-intro}
\end{figure}

\begin{figure*}[t]
\centering
\begin{subfigure}{.49\textwidth}
\centering
\includegraphics[width=\linewidth]{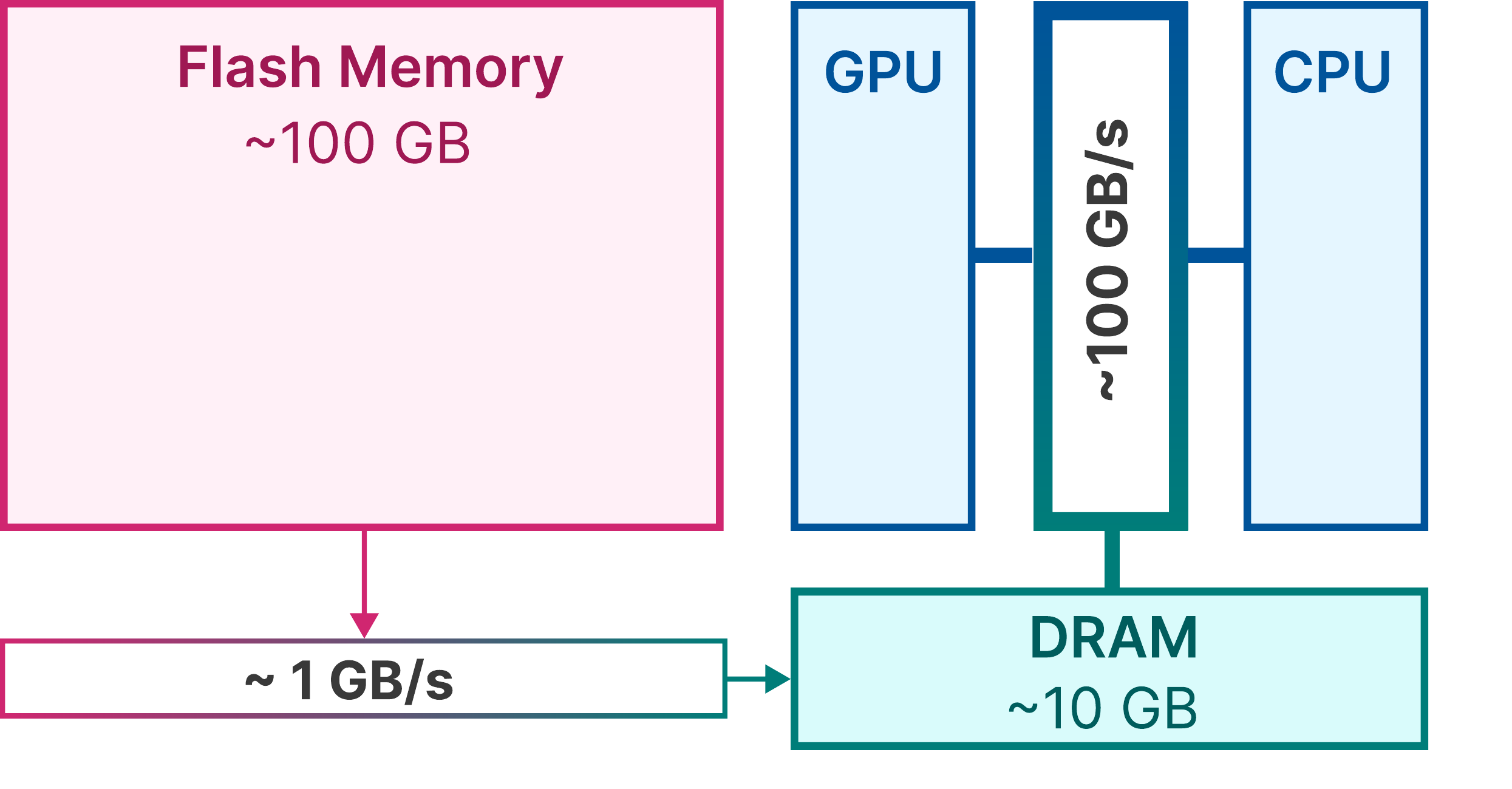}
\caption{Bandwidth in a unified memory architecture}
\label{fig:mem_comp_intro}
\end{subfigure}\hfill
\begin{subfigure}{.41\textwidth}
\centering
\includegraphics[width=\linewidth]{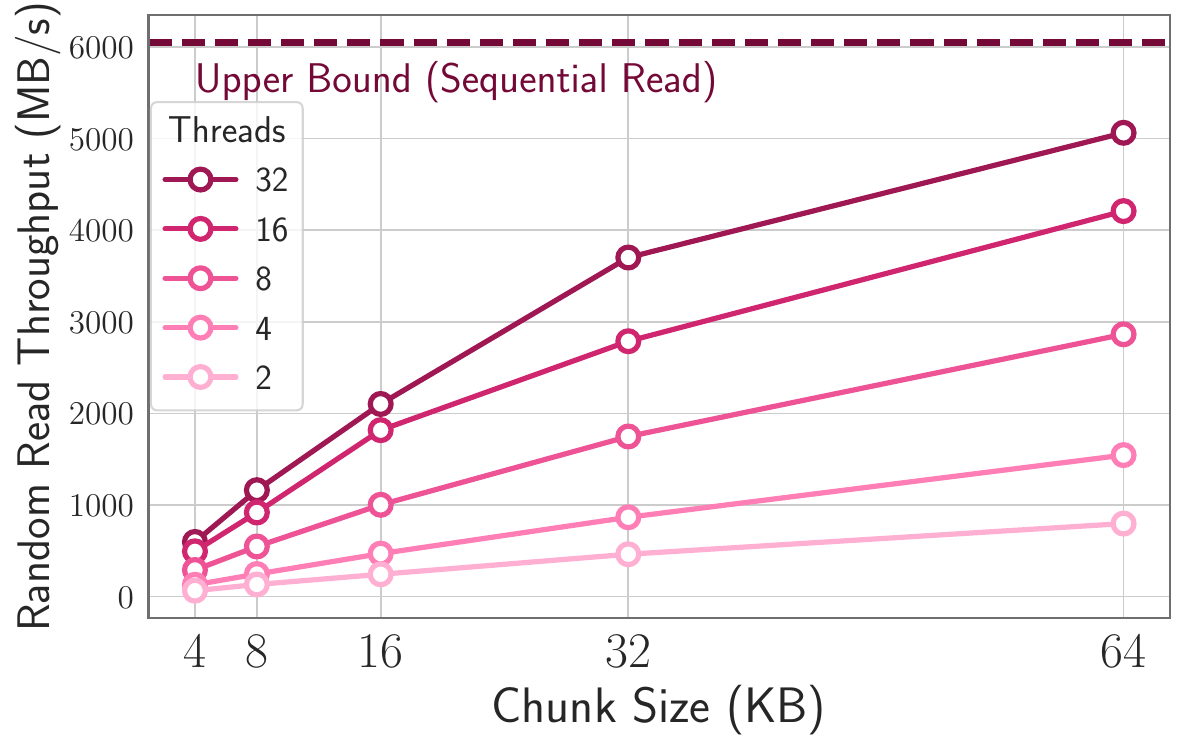}
\caption{Random read throughput of flash memory}
\label{fig:throughput}
\end{subfigure}\hfill
\caption{\textbf{(a)} Flash memory offers significantly higher capacity but suffers from much lower bandwidth compared to DRAM and CPU/GPU caches and registers. \textbf{(b)} The throughput for random reads in flash memory increases with the size of sequential chunks and the number of threads.}
\label{fig:flash_intro}
\end{figure*}

To address this limitation, we propose to store the model parameters in flash memory, which is at least an order of magnitude larger than DRAM. Then, during inference, we directly load the required subset of parameters from the flash memory, avoiding the need to fit the entire model in DRAM. To this end, our work makes several contributions: 
\begin{itemize}[leftmargin=*, noitemsep, topsep=1pt, parsep=0.5pt, partopsep=0pt]%
\item First, we study the hardware characteristics of storage systems (e.g., flash, DRAM). We show that hardware constraints such as capacity and bandwidth limitations can have significant considerations when designing efficient algorithms for serving LLMs from flash (\secref{sec:nand_flash_intro}).
\item Motivated by our findings, we propose several techniques that can help with (i) reducing the required data transfer, (ii) increasing the transfer throughput, and (iii) managing loaded parameters efficiently in DRAM (\secref{sec:load_from_flash}). 
\item Finally, as partially demonstrated in \figref{fig:latency-intro}, we show that our proposed techniques for optimizing the cost model and selectively loading parameters on demand allows us to run models 2x larger than the device's DRAM capacity and speed up inference up to 4x, 7x, and 20x compared to naive implementation in CPU, Metal and NVIDIA GPU backends, respectively (\secref{sec:experiments-and-results}).
\end{itemize}

\section{Flash Memory \& LLM Inference}
\label{sec:nand_flash_intro}

In this section, we explore the characteristics of memory storage systems (e.g., flash, DRAM), and their implications for large language model (LLM) inference. We aim to understand the challenges and hardware-specific considerations essential for algorithm design, particularly in optimizing inference when working with flash memory.

\subsection{Bandwidth and Energy Constraints}
While modern NAND flash memories offer high bandwidth and low latency, they fall well short of the performance levels of DRAM (Dynamic Random-Access Memory), in terms of both latency and throughput. Figure~\ref{fig:mem_comp_intro} illustrates these differences. A naive inference implementation that relies on NAND flash memory might necessitate reloading the entire model for each forward pass. This process is not only time-consuming, often taking seconds for even compressed models, but it also consumes more energy than transferring data from DRAM to the CPU or GPU's internal memory.

Load times for the models can be a problem even in the traditional DRAM-resident setup where weights are not reloaded partially -- the initial, full load of the model still incurs a penalty, particularly in situations requiring rapid response times for the first token. Our approach, leveraging activation sparsity in LLMs, addresses these challenges by enabling selective reading of model weights, thereby reducing the response latency.

\subsection{Read Throughput}
Flash memory systems perform optimally with large sequential reads. For instance, benchmarks on an Apple MacBook M1 Max with 1TB flash memory demonstrate speeds exceeding 6 GiB/s for a 1GiB linear read of an uncached file. However, this high bandwidth cannot be achieved for smaller, random reads due to the inherent multi-phase nature of these reads, encompassing the operating system, drivers, interrupt handling, and the flash controller, among others. Each phase introduces latency, disproportionately affecting smaller reads.

\looseness=-1 To circumvent these limitations, we advocate two primary strategies, which can be employed jointly.
The first involves reading larger chunks of data. For smaller blocks, a substantial part of the overall read time is spent waiting for data transfer to begin. This is often referred to as latency to first byte. This latency reduces the overall throughput of each read operation considerably because the overall measured throughput has to take into account not just the speed of transfer once it begins, but the latency before it begins as well, which penalizes small reads. This means that if we coalesce the reads for rows and columns of the FFN matrices, we can pay the latency cost only once for any given row/column pair in both matrices and higher throughput can be realized. This principle is depicted in \figref{fig:throughput}. Perhaps a counterintuitive yet interesting observation is that in some scenarios, it will be worthwhile to read more than needed (but in larger chunks) and then discard, rather than only reading strictly the necessary parts but in smaller chunks. The second strategy leverages parallelized reads, utilizing the inherent parallelism within storage stacks and flash controllers. Our results indicate that throughputs appropriate for sparse LLM inference are achievable on modern hardware using 32KiB or larger random reads across multiple threads.

Motivated by the challenges described in this section, in \secref{sec:load_from_flash}, we propose methods to optimize data transfer volume and enhance read throughput to significantly enhance inference speeds.

\begin{figure*}[t]
\centering
\begin{subfigure}{.39\textwidth}
  \centering
  \includegraphics[width=\textwidth]{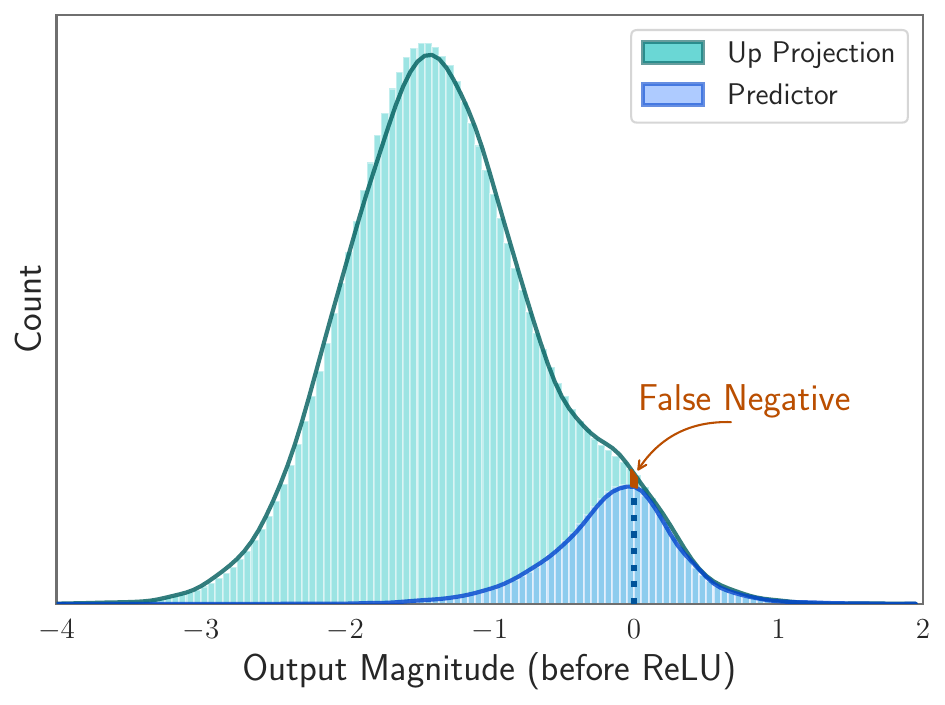}
\caption{predictor vs relu}
  \label{fig:act_compare_sparsity}
\end{subfigure}\hfill
\begin{subfigure}{.48\textwidth}
  \centering
  \includegraphics[width=\textwidth]{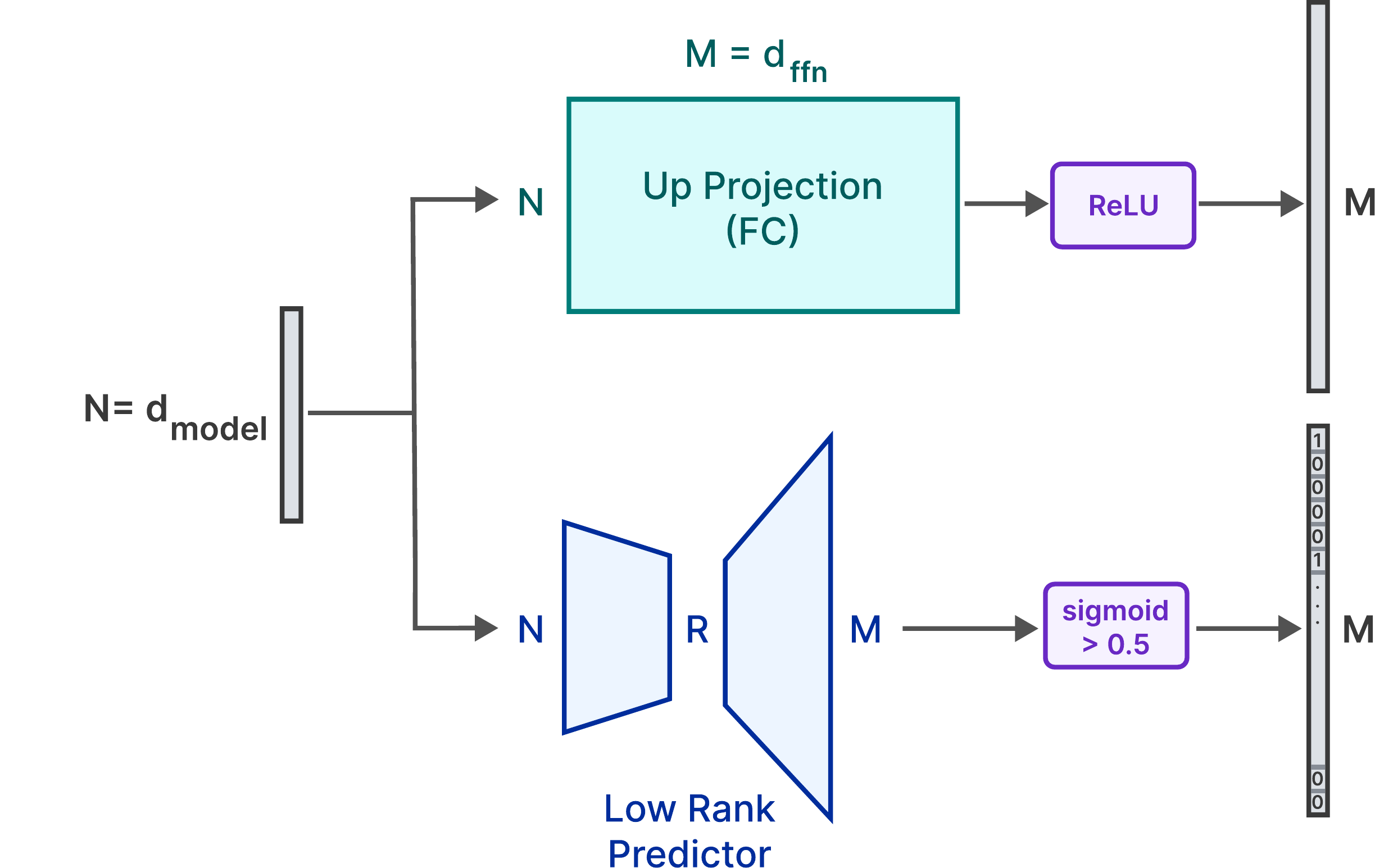}
    \caption{low rank predictor}
  \label{fig:low_rank}
\end{subfigure}\hfill

\caption{\textbf{(a)} Preactivations of tokens in one sequence  in OPT 6.7B. The blue graph shows the preactivation of elements that the predictor detected as positive while the green graph is for up projection. As it can be seen most of the False Positives are close to 0 and False Negatives constitute a small portion of the elements. \textbf{(b)} A small low-rank predictor finds out which intermediate neurons are going to be activated.}
\label{fig:predictor}
\end{figure*}

\section{Load From Flash}
\label{sec:load_from_flash}

This section addresses the challenge of conducting inference on devices where the available DRAM is substantially smaller than the size of the model. This necessitates storing the full model weights in flash memory.
Our primary metric for evaluating various flash loading strategies is latency, dissected into three distinct components: the I/O cost of loading from flash, the overhead of managing memory with newly loaded data, and the compute cost for inference operations.

Our proposed solutions for reducing latency under memory constraints are categorized into  areas:

\begin{enumerate}[leftmargin=*, noitemsep, topsep=1pt, parsep=0.5pt, partopsep=0pt]
    \item \textbf{Reducing Data Load}: Aiming to decrease latency associated with flash I/O operations by loading less data\footnote{It is notable that, by \emph{data} we often refer to the weights of the neural network. However, the techniques we have developed can be easily generalized to other data types transferred and used for LLM inference, such as activations or KV cache, as suggested by~\citet{sheng2023flexgen}.}.
    \item \textbf{Optimizing Data Chunk Size}: Enhancing flash throughput by increasing the size of data chunks loaded, thereby mitigating latency.
    \item \textbf{Efficient Management of Loaded Data}: Streamlining the management of data once it is loaded into memory to minimize overhead.
\end{enumerate}

\looseness=-2 It is important to note that our focus is not on optimizing the compute, as it is orthogonal to the core concerns of our work. Instead, we concentrate on optimizing flash memory interactions and memory management to achieve efficient inference on memory-constrained devices. We will elaborate on the implementation details of these strategies in the experimental setup section.

\subsection{Reducing Data Transfer}

Our method leverages the inherent activation sparsity found in Feed-Forward Network (FFN) models, as documented in preceding research. The OPT 6.7B model, for instance, exhibits a notable 97\% sparsity within its FFN layer. Similarly, the Falcon  7B model has been adapted through fine-tuning, which involves swapping their activation functions to ReLU, resulting in 95\% sparsity while being  similar in accuracy  \citep{mirzadeh2023relu}. Replacing activations of Llama 2 model \citep{touvron2023llama} by FATReLU and finetuning can achieve 90\% sparsity\citep{song2024prosparse}. In light of this information, our approach involves the iterative transfer of only the essential, dynamic subset of the weights from flash memory to DRAM for processing during inference.

\textbf{Selective Persistence Strategy.}
We opt to retain the embeddings and matrices within the attention mechanism of the transformer constantly in DRAM. For the Feed-Forward Network (FFN) portions, only the non-sparse segments are dynamically loaded into DRAM as needed. Keeping attention weights, which constitute approximately one-third of the model's size, in memory, allows for more efficient computation and quicker access, thereby enhancing inference performance without the need for full model loading.

\textbf{Anticipating ReLU Sparsity}.
\looseness=-2 The ReLU activation function naturally induces over 90\% sparsity in the FFN's intermediate outputs, which reduces the memory footprint for subsequent layers that utilize these sparse outputs. However, the preceding layer, namely the up project, must be fully present in memory.

To avoid loading the entire up projection matrix, we follow ~\citet{liu2023deja}, and employ a low-rank predictor to identify the elements zeroed by ReLU (see \figref{fig:low_rank}). We used a balanced loss over negative and positive samples of each layer. In contrast to their work, our predictor needs only the output of the current layer's attention module and not the previous layer's FFN module. We have observed that postponing the prediction to the current layer is sufficient for hardware-aware weight-loading algorithm design but leads to more accurate outcomes due to deferred inputs. We used 10000 samples from the C4 training dataset to do the training for 2 epochs. It took 4 hours on an A100 GPU to train each predictor.

We thereby only load elements indicated by the predictor, as shown in \figref{fig:act_compare_sparsity}. Furthermore, as demonstrated in \tabref{tab:accuracy_metrics}, using predictors does not adversely affect the model's performance in 0-shot tasks. For more details please refer to Appendix \ref{sec:appendix-predictor}.

\begin{table}
\centering
\caption{\label{tab:accuracy_metrics}
The low-rank predictor has a marginal impact on zero-shot metrics as the predictor of each layer accurately identifies sparsity.
}
\resizebox{0.9\linewidth}{!}{
\begin{tabular}{lcc}
\toprule
\textbf{Zero-Shot Task} & \textbf{OPT 6.7B} & \textbf{with Predictor}\\
\midrule
Arc Easy & 66.1 & 66.2 \\
Arc Challenge & 30.6 & 30.6 \\
HellaSwag & 50.3 &  49.8 \\ 
\bottomrule
\end{tabular}}
\vspace{-2mm}
\end{table}

\begin{figure*}[t]
\centering
\begin{subfigure}{.4\textwidth}
  \centering
  \includegraphics[width=\textwidth]{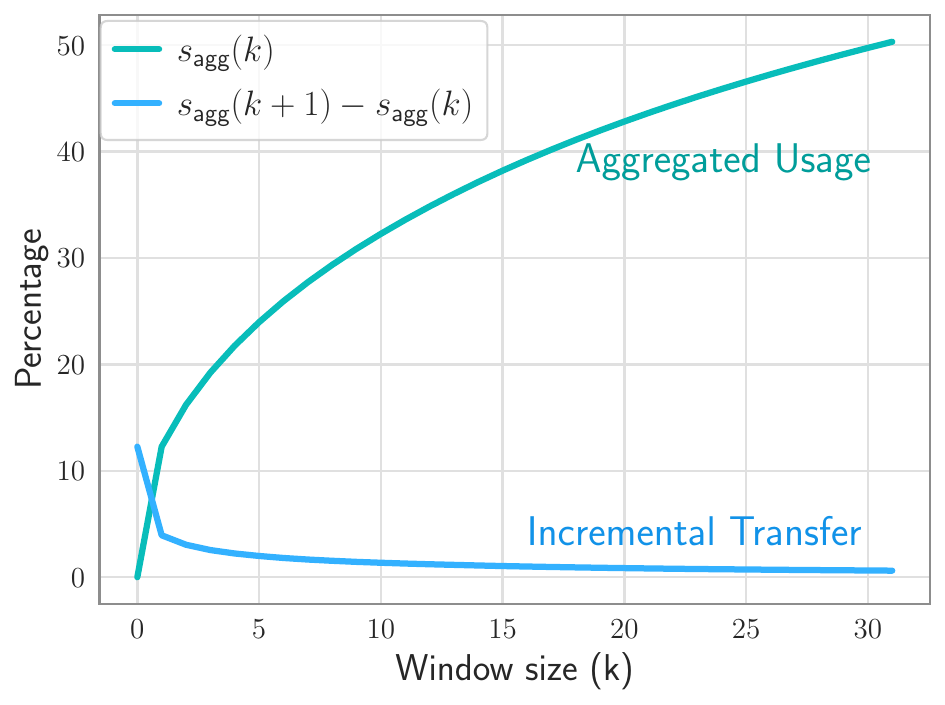}
\caption{aggregated neuron usage}
  \label{fig:aggr_sparsity}
\end{subfigure}\hfill
\begin{subfigure}{.55\textwidth}
  \centering
  \includegraphics[width=\textwidth]{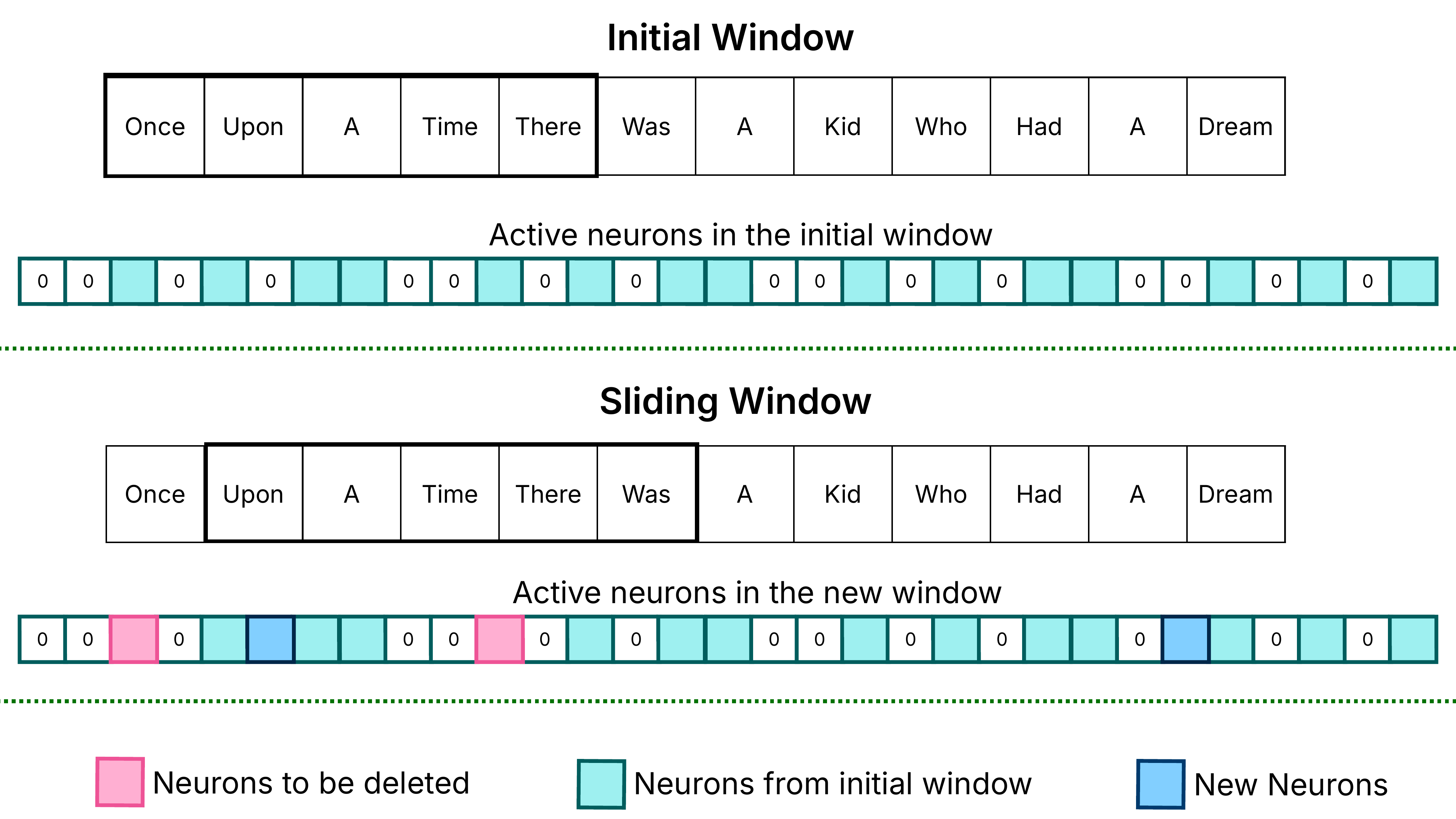}
    \caption{sliding window}
  \label{fig:sliding_window}
\end{subfigure}\hfill

\caption{\textbf{(a)} Aggregated neuron usage of the tenth layer of Falcon 7B: the slope of aggregated neuron usage is decreasing. Other layers exhibit the same pattern. \textbf{(b)} Rather than deleting neurons that were brought to DRAM we keep the active neurons of past $k$ tokens (we use $k=5$): when the new token "Was" is being processed only a small fraction of new weights need to be loaded.}
\label{fig:windowing_aggregation}
\end{figure*}

\textbf{The Sliding Window Technique.}
\looseness=-1 In our study, we define an \textit{active neuron} as one that yields a positive output in our low-rank predictor model. Our approach focuses on managing neuron data by employing a \textit{Sliding Window Technique}. This technique entails maintaining a DRAM cache of only the weight rows that were predicted to be required by the recent subset of input tokens. The key aspect of this technique is the incremental loading of neuron data that differs between the current input token and its immediate predecessors. This strategy allows for efficient memory utilization, as it frees up memory resources previously allocated to cached weights required by tokens that are no longer within the sliding window (as depicted in \figref{fig:sliding_window}).

\looseness=-1 From a mathematical standpoint, let \( s_{\text{agg}}(k) \) denote the cumulative use of neuron data across a sequence of \( k \) input tokens. Our memory architecture is designed to store an average of \( s_{\text{agg}}(k) \) in DRAM. As we process each new token, the incremental neuron data, which is mathematically represented as \( s_{\text{agg}}(k+1) - s_{\text{agg}}(k) \), is loaded from flash memory into DRAM. This practice is grounded in the observed trend of decreasing aggregated neuron usage over time. Consequently, larger values of \( k \) result in a lesser volume of data being loaded for each new token (refer to \figref{fig:aggr_sparsity}), while smaller values of \( k \) can help conserve DRAM that is used to store the cached weights. In determining the size of the sliding window, the aim is to maximize it within the constraints imposed by the available memory capacity.

\subsection{Increasing Transfer Throughput}
\label{sec:increase-chunk-size}
To increase data throughput from flash memory, it is crucial to read data in larger chunks, preferably sized as the multiples of the block size of the underlying storage pool. In this section, we detail the strategy we have employed to augment the chunk sizes for more efficient flash memory reads.

\begin{figure}[t]
  \centering
  \includegraphics[width=0.44\textwidth]{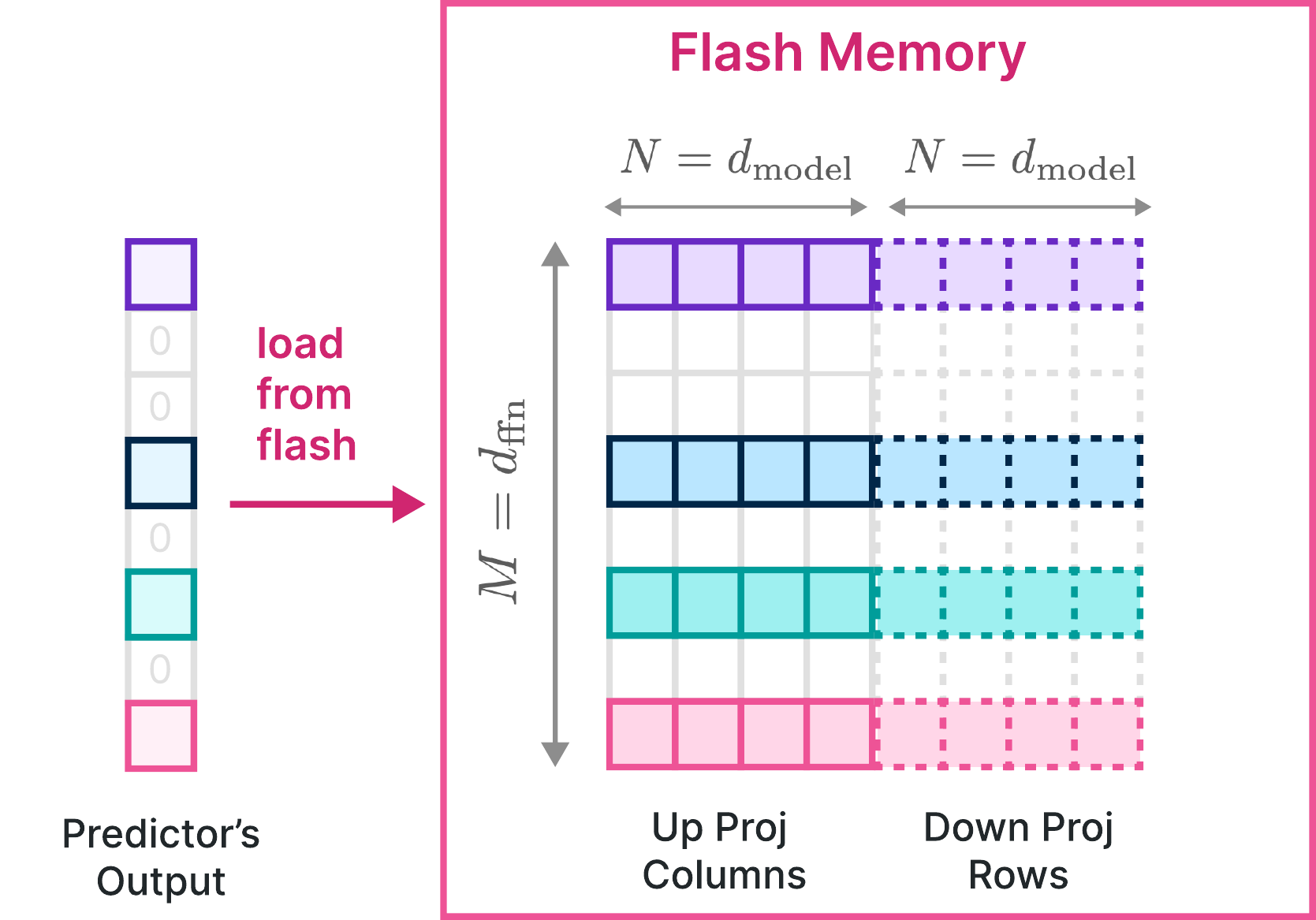}
\caption{ By bundling columns of the up project and rows of the down project layer, we can load 2x chunks instead of reading columns or rows separately.}
\label{fig:row_column_bundling}
\end{figure}

\textbf{Bundling Columns and Rows.}
Note that in the FFN layer, the usage of the $i$th column from the up projection and the $i$th row from the down projection coincides with the activation of the $i$th intermediate neuron. Consequently, by storing these corresponding columns and rows together in flash memory, we can consolidate the data into larger chunks for reading. Refer to \figref{fig:row_column_bundling} for an illustration of this bundling approach. If each element of weights of the network is stored in \emph{num\_bytes} such bundling doubles the chunk size from $d_{model}\times$\emph{num\_bytes} to $2d_{model}\times$\emph{num\_bytes} as shown in \figref{fig:row_column_bundling}. Our analysis and experiment show this increases the throughput of the model.

\textbf{Bundling Based on Co-activation.}
We hypothesized that neurons might exhibit highly correlated activity patterns, enabling bundling. By analyzing activations on the C4 validation dataset, we found a power law distribution of coactivations. However, bundling neurons with their highest coactivated neuron (closest friend) led to multiple loadings of highly active neurons, counteracting our goal. This result suggests that very active neurons are the closest friends of many others. We present this negative result to inspire future research on effective neuron bundling for efficient inference. Please refer to Appendix~\ref{sec:appendix-bundling-coactivation} for details.

\subsection{Optimized Data Management in DRAM}

\begin{figure*}[t]
\centering
\includegraphics[width=0.9\textwidth]{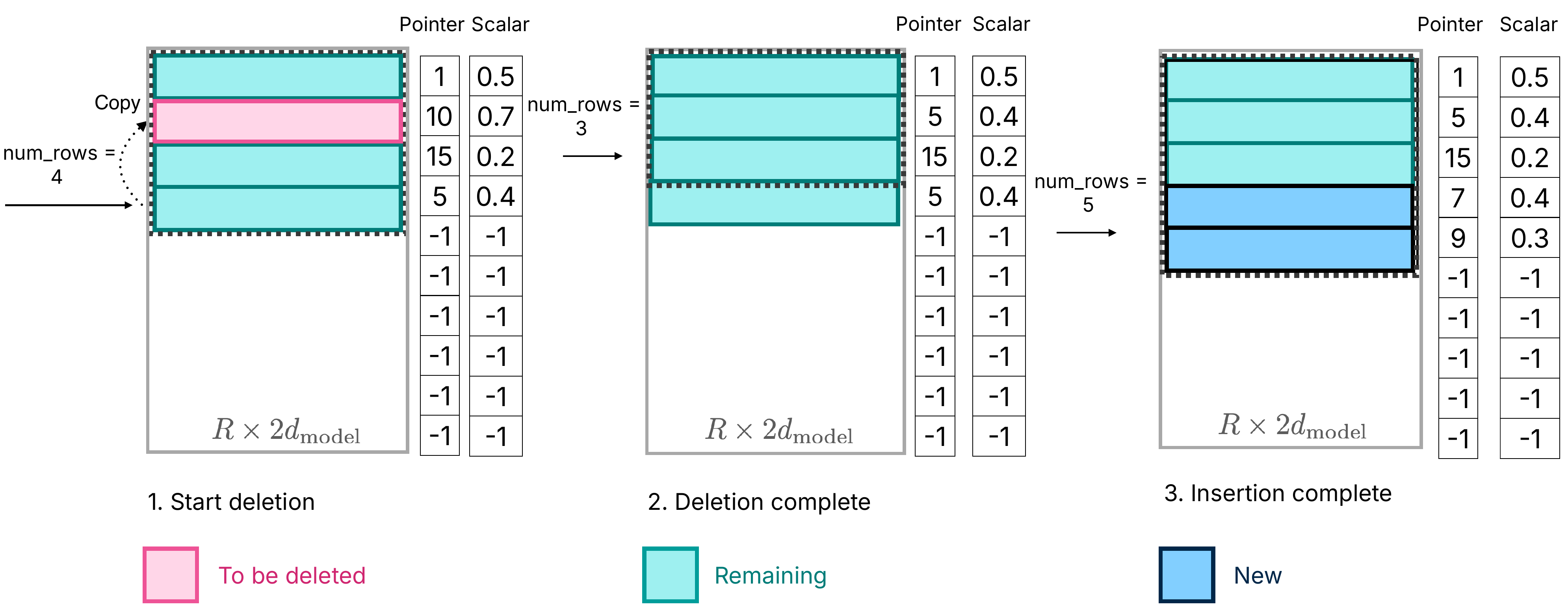}
\caption{ Memory management; First we replace elements to be deleted by last elements to maintain a consecutive occupation of memory. Then the new weights are stacked at the end. This reduces the unnecessary data movements.}
\label{fig:memory_management}
\end{figure*}

\looseness=-1 Although data transfer within DRAM is more efficient compared to accessing flash memory, it still incurs a non-negligible cost. When introducing data for new neurons, reallocating the matrix and appending new matrices can lead to significant overhead due to the need for rewriting existing neuron data in DRAM. This is particularly costly when a substantial portion (approximately 25\%) of the Feed-Forward Networks (FFNs) in DRAM needs to be rewritten. To address this issue, we adopt an alternative memory management strategy. This involves the preallocation of all necessary memory and the establishment of a corresponding data structure for efficient management. The data structure comprises elements such as \texttt{pointers}, \texttt{matrix}, \texttt{bias}, \texttt{num\_used}, and \texttt{last\_k\_active} shown in \figref{fig:memory_management}.

Each row in the \texttt{matrix} represents the concatenated row of the `up project' and the column of the `down project' of a neuron. The \texttt{pointer} vector indicates the original neuron index corresponding to each row in the \texttt{matrix}. The bias for the `up project' in the original model is represented in the corresponding \texttt{bias} element. The \texttt{num\_used} parameter tracks the number of rows currently utilized in the \texttt{matrix}, initially set to zero. The \texttt{matrix} for the $i$th layer is pre-allocated with a size of $\texttt{Req}_i \times 2d_{\text{model}}$, where $\texttt{Req}_i$ denotes the maximum number of neurons required for the specified window size in a subset of C4 validation set. By allocating enough memory for each layer in advance, we minimize the need for reallocation. Finally, the \texttt{last\_k\_active} component identifies the neurons from the original model that were most recently activated using the last $k$ tokens. The following operations can be done as depicted in \figref{fig:memory_management}:

\begin{enumerate}[leftmargin=*, noitemsep, topsep=0pt, parsep=0pt, partopsep=0pt]
    \item \textbf{Deleting Neurons:} Neurons that are no longer required are identified efficiently in linear time, utilizing the associated \texttt{last\_k\_active} value and the current prediction. The \texttt{matrix}, \texttt{pointer}, and \texttt{scalars} of these redundant neurons are replaced with the most recent elements, and their count is subtracted from \texttt{num\_rows}. For $O(c)$ neurons to be deleted, a memory rewrite of the order $O(c \times d_{\text{model}})$ is required.

    \item \textbf{Bringing in New Neurons:} The required weights are retrieved from flash memory. The corresponding pointers and scalars are read from DRAM, and these rows are then inserted into the matrix, extending from \texttt{num\_row} to \texttt{num\_row + num\_new}. This approach eliminates the need for reallocating memory in DRAM and copying existing data, reducing inference latency.

    \item \textbf{Inference Process:} For the inference operation, the first half of the \texttt{matrix[:num\_rows,:d\_model]} is used as the `up project', and the transposed second half, \texttt{matrix[:num\_rows,d\_model:].transpose()}, serves as the 'down project'. This configuration is possible because the order of neurons in the intermediate output of the FFN does not alter the final output, allowing for a streamlined inference process.
\end{enumerate}

These steps collectively ensure efficient memory management during inference, optimizing the neural network's performance and resource utilization.

\section{Experiments and Results}
\label{sec:experiments-and-results}
\looseness=-2 We start this section by briefly discussing our experimental setup and implementation details. Next, we show that the techniques introduced in \secref{sec:load_from_flash} can improve the inference latency significantly across different models and runtime platforms. We postpone the some details to the appendix sections as follows: performance of our trained low-rank predictor (Appendix \ref{sec:appendix-predictor}).

\subsection{Experimental Setup}
Our work is mainly motivated by optimizing inference efficiency on personal devices. To this end, in our experiments, we process sequences individually, running only one sequence at a time. This approach allows us to allocate a specific portion of DRAM for the Key-Value (KV) cache while primarily focusing on the model size. For the implementation of our inference process, we utilize HuggingFace Transformers library~\citep{huggingface_transformers} and PyTorch~\citep{pytorch}. This setup is tested under the condition that approximately half of the model size is available in DRAM. While with a different level of sparsity or employing quantization, one can work with smaller available DRAM capacity, these optimizations are orthogonal to our proposed method.

\begin{table*}[t]
\centering
\caption{The I/O latency of OPT 6.7B 16 bit on M1 Max when half the memory is available. By employing the activation predictor and windowing, we can reduce the data transfer from flash memory to DRAM. While this reduces the throughput, the bundling technique can alleviate this by doubling the data transfer chunk size and hence the throughput which leads to reducing the overall latency to half.} 
\label{tab:IO_per_method}
\resizebox{0.98\textwidth}{!}{%
\begin{tabular}{@{}c|ccc|cccc@{}}
\toprule
\multicolumn{4}{c|}{\textbf{Configuration}} &
  \multicolumn{4}{c}{\textbf{Performance Metrics}} \\ \midrule
\textbf{Hybrid} &
  \textbf{Predictor} &
  \textbf{Windowing} &
  \textbf{Bundling} &
  \multicolumn{1}{c|}{\textbf{DRAM (GB)}} &
  \multicolumn{1}{c|}{\textbf{Flash$\rightarrow$ DRAM (GB)}} &
  \multicolumn{1}{c|}{\textbf{Throughput (GB/s)}} &
  \textbf{I/O Latency (ms)} \\ \midrule
  \xmark &
  \xmark &
  \xmark &
  \xmark &
  \multicolumn{1}{c|}{0} &
  \multicolumn{1}{c|}{13.4 GB} &
  \multicolumn{1}{c|}{6.10 GB/s} &
  2196 ms \\
  \cmark &
  \xmark &
  \xmark &
  \xmark &
  \multicolumn{1}{c|}{6.7} &
  \multicolumn{1}{c|}{6.7 GB} &
  \multicolumn{1}{c|}{6.10 GB/s} &
  1090 ms \\
  \cmark &
  \cmark &
  \xmark &
  \xmark &
  \multicolumn{1}{c|}{4.8} &
  \multicolumn{1}{c|}{0.9 GB} &
  \multicolumn{1}{c|}{1.25 GB/s} &
  738 ms \\
  \cmark &
  \cmark &
  \cmark &
  \xmark &
  \multicolumn{1}{c|}{6.5} &
  \multicolumn{1}{c|}{0.2 GB} &
  \multicolumn{1}{c|}{1.25 GB/s} &
  164 ms \\
  \cmark &
  \cmark &
  \cmark &
  \cmark &
  \multicolumn{1}{c|}{6.5} &
  \multicolumn{1}{c|}{0.2 GB} &
  \multicolumn{1}{c|}{2.25 GB/s} &
  87 ms \\ \bottomrule
\end{tabular}%
}
\end{table*}

\textbf{Models.}
We mainly use OPT 6.7B~\citep{OPTpaper} and the sparsified Falcon 7B~\citep{mirzadeh2023relu} model for our evaluations, but we additionally report results on Phi-2~\citep{phi_paper}, Persimmon 8B~\citep{persimmon-8b} and a Llama 2~\citep{touvron2023llama} which is sparsified using FATReLU ~\citep{song2024prosparse}. Note that the techniques introduced in this work are mostly independent of architecture.

\textbf{Data.} We use a small subset of C4 validation dataset for our latency measurements. We take the first 128 tokens of each example as the prompt, and generate 256 new tokens. 

\textbf{Hardware Configuration.}
Our models are evaluated across three hardware setups. The first includes an Apple M1 Max with a 1TB SSD. The second features an Apple M2 Ultra with a 2TB SSD. On MacBooks we run the model on the CPU with float32 or GPU with Metal and float16. The third setup uses a Linux machine with a 24GB NVIDIA RTX 4090, where GPU computations utilize bfloat16 models. Across all setups, we assume nearly half of the total memory (DRAM and GPU) is allocated for model computations.

\textbf{Baselines.}
\looseness=-1 We compare our models with a \emph{naive} baseline of loading the model on demand when doing the forward pass. We additionally compare with our \emph{hybrid} loading approach as a secondary baseline when half of the model is persisted in memory and the other half is loaded on demand at generation of every token without use of sparsity. We used best theoretical possible numbers for IO latency for each of the methods to make a fair comparison, the real number might be higher.
\looseness=-1 For methods not employing sparsity or weight sharing, at least half of the model must be transferred from flash memory during the forward pass. This necessity arises because, initially, only half of the model is available in DRAM, but as the forward pass progresses, the entire model capacity is utilized. Consequently, any data not present at the start must be transferred at least once. Thus, the most efficient theoretical baseline involves loading half of the model size from the flash memory into DRAM. This optimal I/O scenario serves as our primary baseline. Given the nature of our setup (i.e., the limited available DRAM or GPU memory), we are not aware of any other method that can surpass this theoretical I/O efficiency.

\textbf{Implementation Details.}
\looseness=-1 To optimize data loading from flash memory, our system employs reads parallelized over 32 threads. This multithreaded approach is intended to both better amortize latency to the first byte by not waiting for each read sequentially, and maximize read throughput by reading multiple streams at once (\figref{fig:throughput}). To better assess the actual throughput, we conducted benchmarks without the aid of operating system caching leading to a more accurate measurement.

\vspace{-1mm}
\subsection{Faster Load From Flash}
Our first result in \tabref{tab:IO_per_method} demonstrates the effectiveness of techniques we introduced in \secref{sec:load_from_flash}, where the I/O latency depends on how much data is being transferred from flash to DRAM, and the chunk size which determines the throughput. For instance, by using a low-rank predictor, we reduce the data transfer significantly, and the amount of this traffic can be further reduced using our proposed windowing technique. Compared to a long, contiguous read, scattered reads will necessarily result in lower throughput (e.g. 1.25 GiB/s sparse vs 6.1 GiB/s dense), but this is partially mitigated by bundling up-projection and down-projection weights. The overall effect of sparse reads is still strongly favorable, because only a small subset of the overall weights is loaded incrementally in each iteration, and the load of just the required subset of weights takes less time and less DRAM.

\begin{table}[t]
\centering
\caption{The end-to-end inference latency across different setups. Our efficient implementation (referred as \emph{All}) that employs the predictor, windowing, and bundling can lead to significant latency reduction.}
\label{tab:full-e2e-results}
\resizebox{0.495\textwidth}{!}{%
\begin{tabular}{@{}lll|cccc@{}}
\cmidrule(l){4-7}
                                    &    &     & \multicolumn{4}{c}{\textbf{Inference Latency (ms)}}                                            \\ \midrule
\multicolumn{1}{l|}{\textbf{Model}} & \textbf{Method} & \textbf{Backend} & \textbf{I/O} & \textbf{Mem} & \textbf{Compute} & \textbf{Total} \\ \midrule
\multicolumn{1}{l|}{OPT 6.7B}       & Naive & CPU              & 2196                     & 0                          & 986              & 3182           \\
\multicolumn{1}{l|}{OPT 6.7B}       & All & CPU       & \textbf{105}                      & 58                         & 506              & \textbf{669}            \\ \hdashline
\multicolumn{1}{l|}{OPT 6.7B}       & Naive & Metal M1              & 2196                     & 0                          & 193              & 2389           \\
\multicolumn{1}{l|}{OPT 6.7B}       & All & Metal M1       & \textbf{92}                      & 35                         & 438              & \textbf{565}            \\ \hdashline
\multicolumn{1}{l|}{OPT 6.7B}       & Naive & Metal M2              &      2145                & 0                          & 125              & 2270           \\
\multicolumn{1}{l|}{OPT 6.7B}       & All & Metal M2       & \textbf{26}                      & 8                         & 271              & \textbf{305}            \\ \hdashline

\multicolumn{1}{l|}{OPT 6.7B}       & Naive & GPU              & 2196                     & 0                          & 22               & 2218           \\ 

\multicolumn{1}{l|}{OPT 6.7B}       & All & GPU        & \textbf{30}                       & 34                         & 20               & 84             \\ 
\multicolumn{1}{l|}{OPT 6.7B}       & Speculative & GPU        & 38.5                       & 9.5                         & 12               & \textbf{60}             \\ \midrule

\multicolumn{1}{l|}{Falcon 7B}      & Naive & CPU              & 2295                     & 0                          & 800              & 3095           \\
\multicolumn{1}{l|}{Falcon 7B}      & Hybrid & CPU              & 1147                     & 0                          & 800              & 1947           \\

\multicolumn{1}{l|}{Falcon 7B}      & All & CPU        & \textbf{161}                      & 92                         & 453              & \textbf{706}            \\ \midrule
\multicolumn{1}{l|}{Persimmon 8B}   & Naive & CPU              & 2622                     & 0                          & 1184             & 3806           \\
\multicolumn{1}{l|}{Persimmon 8B}   & Hybrid & CPU              & 1311                     & 0                          & 1184             & 2495           \\

\multicolumn{1}{l|}{Persimmon 8B}   & All & CPU        & \textbf{283}                      & 98                         & 660              & \textbf{1041}           \\ \midrule
\multicolumn{1}{l|}{Phi-2 2.7B}   & Naive & CPU              & 885                      & 0                          & 402              & 1287           \\
\multicolumn{1}{l|}{Phi-2 2.7B}  & Hybrid & CPU              & 309                      & 0                          & 402              & 711           \\

\multicolumn{1}{l|}{Phi-2 2.7B}          & All & CPU       & \textbf{211}                      & 76                         & 259              & \textbf{546}            \\ \midrule
\multicolumn{1}{l|}{Llama 2 7B}   & Naive & CPU              & 2166                      & 0                          & 929              & 3095           \\
\multicolumn{1}{l|}{Llama 2 7B}  & Hybrid & CPU              & 974                      & 0                          & 929              & 1903           \\

\multicolumn{1}{l|}{Llama 2 7B}          & All & CPU       & \textbf{279}                      & 152                         & 563              & \textbf{994}            \\ \bottomrule
\end{tabular}%
}
\vspace{-1mm}
\end{table}

Additionally, we examine end-to-end latencies under various setups in \tabref{tab:full-e2e-results}. We allocate approximately 50 \% of the model size for OPT, Falcon, and Persimmon and Llama 2. For the significantly smaller Phi-2 model, we observed less sparsity rates, prompting us to set this limit at 65\%. We observe a significant improvement in loading efficiency over both naive and hybrid approaches across all models. Moreover we showed the GPU backend outcomes further improve when combined with speculative decoding. %

\subsection{The Memory-Latency Tradeoff}
So far, we have mainly worked under the assumption that the available DRAM is roughly half of our model size. However, we note that this is not a hard constraint and we can relax this constraint.

\begin{figure}[t]
  \centering
  \includegraphics[width=0.42\textwidth]{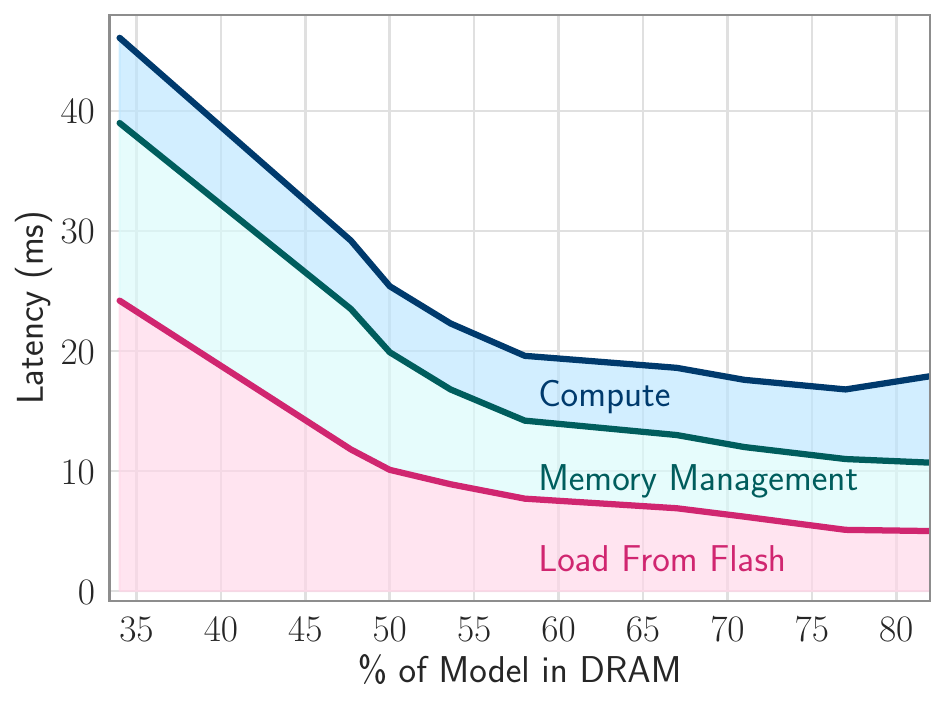}
\caption{By bringing more of our model (OPT-6.7B) parameters into DRAM, the latency can be reduced on the GPU machine.}
\label{fig:mem_latency_tradeoff}
\vspace{-2mm}
\end{figure}

To this end, we study the impact of window size on memory usage, and consequently on latency. By increasing the window size, we increase the percentage of model parameters that we keep in DRAM. As a result, we need to bring fewer parameters, and hence the latency can be reduced at the cost of using higher DRAM as shown in \figref{fig:mem_latency_tradeoff}.

\section{Ablation analysis}
\label{sec:ablation-analysis}

\subsection{The Impact of Longer Generation} %
\begin{figure}[t]
  \centering
  \includegraphics[width=0.42\textwidth]{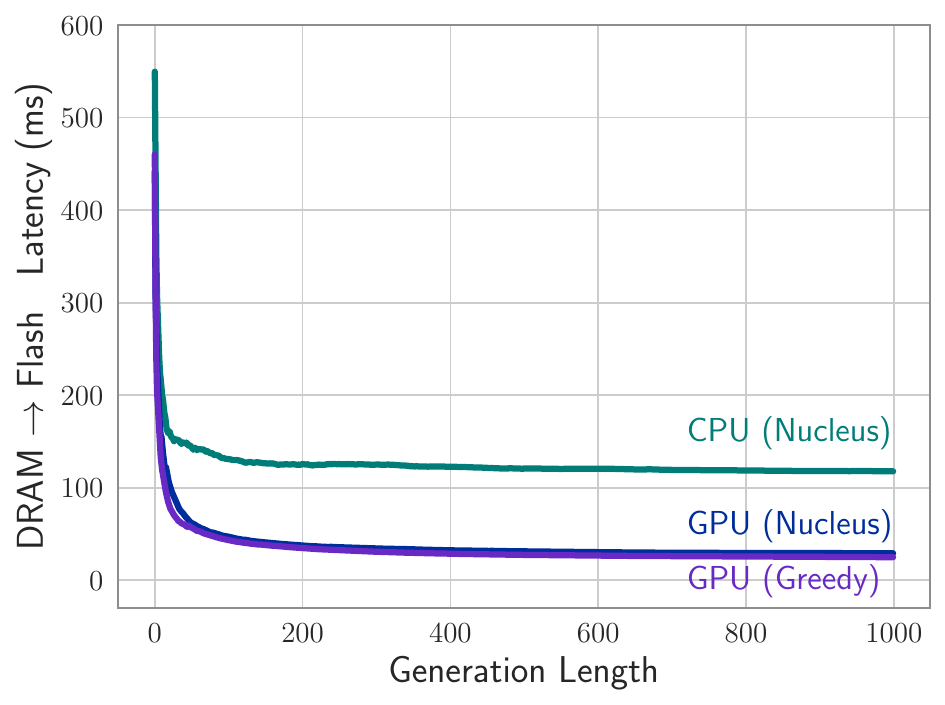}
\caption{Weight loading latency of OPT 6.7B with increasing generation length.}
\label{fig:long_generation_nucleus}
\end{figure}

In our previous results, we have used short to medium-length (256 tokens) generations for our benchmarks. It is possible that for longer generation of tokens, the ssd enable thermal throttling and lower the performance. However, \figref{fig:long_generation_nucleus} shows that this is not the case, even when we generate 1000 tokens for OPT 6.7B model on GPU.  Moreover, we show that the average flash latency doesn't increase as we go further in generation. In contrast, the flash latency for the the first few tokens is higher since the allocated memory in DRAM is empty and needs to be filled in with neurons and for first few tokens we need more data transfer.

Also it is possible to argue that the non-greedy sampling methods such as the Nucleus sampling~\citep{Nucleus_Sampling} method can result in more diverse activation, and hence less favorable towards our method. We found out this is not the case either for long token generations. Nucleus sampling doesn't lead to lower performance in long generation in neither cpu or gpu.

\subsection{Speculative Decoding} %
To further showcase the strength of our method and adaptability to other decoding strategies we have applied speculative decoding on the OPT 6.7B model. The challenge for doing speculative decoding is the limited memory available within DRAM. Given $\lambda$ tokens from the draft model, the big model verifies them and will keep a window of size $k$ for each layer. The model should decide neurons of which tokens to keep in memory before verifications are done. If the model keeps the last $k$ tokens out of $\lambda+1$ tokens in memory and most of them get rejected, there will be very few neuron reuse for the next forward pass. We conjecture that if the ratio of the acceptance is $\alpha$ keeping the last $k$ tokens ending with $\alpha (\lambda+1)$th token is optimal in DRAM. We used $\lambda=4$ and were able to improve the speed of decoding by 1.4x as shown in table \ref{tab:full-e2e-results-with-std}, this is close to the original 1.58x speedup of speculative decoding.

\subsection{A Note on Power Consumption}
In evaluating the efficiency of our method, we compared the power consumption of our sparse model approach with that of generating tokens using a dense model of similar size. While the power usage (energy per unit of time) of the sparse model was lower than that of the dense model, the extended duration required for token generation resulted in the sparse model having a higher total energy consumption. This is going to be reflected in the greater area under the curve when plotting power over time for the sparse model compared to the dense model. A systematic and quantitative evaluation of the exact power usage pattern is left as a future work.

\section{Related Works}
\label{sec:related-work}
\looseness=-1 As LLMs grow in size, reducing their computational and memory requirements for inference has become an active area of research. Approaches broadly fall into two categories: model compression techniques such as pruning and quantization \citep{han2015deep, Sun2023ASA, Jaiswal2023CompressingLT, Xia2023FlashLLMEL, zhang2022llm, Xu2023CompressTP, Shao2023OmniQuantOC, Lin2023AWQAW, Hoang2023DynamicPM, Zhao2023AtomLQ, Ahmadian2023IntriguingPO, Li2023NormTH}, and selective execution such as sparse activation \citep{liu2023deja, mirzadeh2023relu}, or conditional computation \citep{graves2016adaptive, Baykal2023AlternatingUF}. Our work is orthogonal to these directions, focusing mainly on minimizing data transfer from flash memory during inference.

\looseness=-2 Perhaps most related to our work is the literature on selective weight loading. Dejavu \citep{liu2023deja} exploits activation sparsity to load a subset of weights for each layer. However, it still requires loading from GPU memory. FlexGen \citep{sheng2023flexgen} offloads the weights and KV-cache from GPU memory to DRAM and DRAM to flash memory. In contrast, we consider only the cases where the full model can't reside in the whole DRAM and GPU memory on the edge devices. Notably, FlexGen is still theoretically bound by the slow throughput of flash to DRAM in such scenarios. An expanded discusion of related works is deferred to Appendix~\ref{sec:appendix-related-works}.

Overall, the primary assumption in the literature is that the model can fully reside in the GPU memory or system DRAM. However, considering the limited resources available on personal devices, we do not share this assumption in this work. Instead, we concentrate on exploring how to store and load parameters on flash memory more efficiently, aiming to enhance inference efficiency.

\section{Discussion}
In this study, we have tackled the significant challenge of running large language models (LLMs) on devices with constrained memory capacities. Our approach, deeply rooted in the understanding of flash memory and DRAM characteristics, represents a novel convergence of hardware-aware strategies and machine learning. By developing an inference cost model that aligns with these hardware constraints, we have introduced two new techniques: `windowing' and `row-column bundling'.

\looseness=-1 The practical outcomes of our research are noteworthy. We have demonstrated the ability to run LLMs up to twice the size of available DRAM. For example, on OPT model, we achieve an acceleration in inference speed of 4-5x compared to traditional loading methods in CPU, and 20-25x in GPU. This is particularly crucial for deploying LLMs in resource-limited environments, thereby expanding their applicability and accessibility.

\looseness=-1 While in this work we have studied the previously unexplored problem of serving LLMs from flash, we note that this work is only a first step in this direction, and has several limitations that we discuss in the next section, and we believe there are several interesting problems left to be explored in future works. For instance, from the algorithmic perspective, more optimized techniques of weight bundling and data structures can be crafted, while from the engineering perspective, the characteristics of specific hardware platforms can inform works on building more efficient inference stacks. %

\newpage \clearpage
\section{Limitations}
Our study represents an initial endeavor in the pursuit of democratizing Large Language Model (LLM) inference, making it accessible to a wider array of individuals and devices. We recognize that this early effort has its limitations, which, in turn, open up compelling avenues for future research. 
A critical aspect for future exploration is the systematic analysis of power consumption and thermal limitations inherent in the methods we propose, particularly for on-device deployment. 

Currently, our study is limited to single-batch inference. We provide some preliminary results on combining our proposed idea with speculative decoding, however, expanding this to include more complex scenarios like prompt processing and multi-batch inference are valuable areas for further investigation. 

In our initial proof of concept, we operated under the assumption of memory availability being half the size of the model. Exploring the dynamics of working with varying memory sizes—both larger and smaller—introduces a fascinating balance between latency and accuracy, and is a compelling area for future exploration. 

In conclusion, our methodology is constructed on the foundation of sparsified networks. Nonetheless, the underlying concept holds potential for broader applications. It can be adapted to selectively load weights in non-sparse networks or to dynamically retrieve model weights from flash storage. This adaptation would be contingent on the specific requirements of the input prompt or the contextual parameters provided. Such an approach suggests a versatile strategy for managing model weights, and optimizing performance based on the nature of the input, thereby enhancing the efficiency, usefulness, and applicability of the proposed scheme in various scenarios dealing with Large Language Models (LLMs).

\section*{Acknowledgements}
We would like to thank Itay Sagron, Lailin Chen, Chenfan (Frank) Sun, Hanieh Hashemi, Mahyar Najibi, Qichen Fu, Moin Nabi, Peter Zatloukal, Arsalan Farooq, Sachin Mehta, Mohammad Samragh,  Matt Johnson, Etai Zaltsman, Lin Chang, Dominic Giampaolo,  Tal Uliel, Hadi Pouransari, Fartash Faghri, Oncel Tuzel, Samy Bengio, Ruoming Pang, Chong Wang, Ronan Collobert, David Grangier, and Aftab Munshi for the valuable  discussions.

\bibliography{refs}

\begin{thebibliography}{54}
\expandafter\ifx\csname natexlab\endcsname\relax\def\natexlab#1{#1}\fi

\bibitem[{Ahmadian et~al.(2023)Ahmadian, Dash, Chen, Venkitesh, Gou, Blunsom,
  Ustun, and Hooker}]{Ahmadian2023IntriguingPO}
Arash Ahmadian, Saurabh Dash, Hongyu Chen, Bharat Venkitesh, Stephen Gou, Phil
  Blunsom, A.~Ustun, and Sara Hooker. 2023.
\newblock \href {https://api.semanticscholar.org/CorpusID:258967189}
  {Intriguing properties of quantization at scale}.
\newblock \emph{ArXiv}, abs/2305.19268.

\bibitem[{Almazrouei et~al.(2023)Almazrouei, Alobeidli, Alshamsi, Cappelli,
  Cojocaru, Alhammadi, Daniele, Heslow, Launay, Malartic, Noune, Pannier, and
  Penedo}]{FalconPaper}
Ebtesam Almazrouei, Hamza Alobeidli, Abdulaziz Alshamsi, Alessandro Cappelli,
  Ruxandra Cojocaru, Maitha Alhammadi, Mazzotta Daniele, Daniel Heslow, Julien
  Launay, Quentin Malartic, Badreddine Noune, Baptiste Pannier, and Guilherme
  Penedo. 2023.
\newblock The falcon series of language models: Towards open frontier models.

\bibitem[{Aminabadi et~al.(2022)Aminabadi, Rajbhandari, Awan, Li, Li, Zheng,
  Ruwase, Smith, Zhang, Rasley et~al.}]{aminabadi2022deepspeed}
Reza~Yazdani Aminabadi, Samyam Rajbhandari, Ammar~Ahmad Awan, Cheng Li, Du~Li,
  Elton Zheng, Olatunji Ruwase, Shaden Smith, Minjia Zhang, Jeff Rasley, et~al.
  2022.
\newblock Deepspeed-inference: enabling efficient inference of transformer
  models at unprecedented scale.
\newblock In \emph{SC22: International Conference for High Performance
  Computing, Networking, Storage and Analysis}, pages 1--15. IEEE.

\bibitem[{Bae et~al.(2023)Bae, Ko, Song, and Yun}]{Bae2023FastAR}
Sangmin Bae, Jongwoo Ko, Hwanjun Song, and Se-Young Yun. 2023.
\newblock \href {https://api.semanticscholar.org/CorpusID:263830054} {Fast and
  robust early-exiting framework for autoregressive language models with
  synchronized parallel decoding}.
\newblock \emph{ArXiv}, abs/2310.05424.

\bibitem[{Baykal et~al.(2023)Baykal, Cutler, Dikkala, Ghosh, Panigrahy, and
  Wang}]{Baykal2023AlternatingUF}
Cenk Baykal, Dylan Cutler, Nishanth Dikkala, Nikhil Ghosh, Rina Panigrahy, and
  Xin Wang. 2023.
\newblock \href {https://api.semanticscholar.org/CorpusID:256416125}
  {Alternating updates for efficient transformers}.
\newblock \emph{ArXiv}, abs/2301.13310.

\bibitem[{Brown et~al.(2020)Brown, Mann, Ryder, Subbiah, Kaplan, Dhariwal,
  Neelakantan, Shyam, Sastry, Askell et~al.}]{brown2020language}
Tom Brown, Benjamin Mann, Nick Ryder, Melanie Subbiah, Jared~D Kaplan, Prafulla
  Dhariwal, Arvind Neelakantan, Pranav Shyam, Girish Sastry, Amanda Askell,
  et~al. 2020.
\newblock Language models are few-shot learners.
\newblock \emph{Advances in neural information processing systems},
  33:1877--1901.

\bibitem[{Chia et~al.(2023)Chia, Hong, Bing, and Poria}]{chia2023instructeval}
Yew~Ken Chia, Pengfei Hong, Lidong Bing, and Soujanya Poria. 2023.
\newblock \href {http://arxiv.org/abs/2306.04757} {Instructeval: Towards
  holistic evaluation of instruction-tuned large language models}.

\bibitem[{Chowdhery et~al.(2022)Chowdhery, Narang, Devlin, Bosma, Mishra,
  Roberts, Barham, Chung, Sutton, Gehrmann et~al.}]{chowdhery2022palm}
Aakanksha Chowdhery, Sharan Narang, Jacob Devlin, Maarten Bosma, Gaurav Mishra,
  Adam Roberts, Paul Barham, Hyung~Won Chung, Charles Sutton, Sebastian
  Gehrmann, et~al. 2022.
\newblock Palm: Scaling language modeling with pathways.
\newblock \emph{arXiv preprint arXiv:2204.02311}.

\bibitem[{Dai et~al.(2021)Dai, Zhang, Gong, Yang, Dai, Song, and
  Xie}]{dai2021spatten}
Han Dai, Yi~Zhang, Ziyu Gong, Nanqing Yang, Wei Dai, Eric Song, and Qiankun
  Xie. 2021.
\newblock Spatten: Efficient sparse attention architecture with cascade token
  and head pruning.
\newblock In \emph{Advances in Neural Information Processing Systems},
  volume~34.

\bibitem[{Elsen et~al.(2023)Elsen, Odena, Nye, Ta\c{s}\i{}rlar, Dao, Hawthorne,
  Moparthi, and Somani}]{persimmon-8b}
Erich Elsen, Augustus Odena, Maxwell Nye, Sa\u{g}nak Ta\c{s}\i{}rlar, Tri Dao,
  Curtis Hawthorne, Deepak Moparthi, and Arushi Somani. 2023.
\newblock \href {https://www.adept.ai/blog/persimmon-8b} {Releasing
  {Persimmon-8B}}.

\bibitem[{Gale et~al.(2020)Gale, Zaharia, Young, and Elsen}]{gale2020sparse}
Trevor Gale, Matei Zaharia, Cliff Young, and Erich Elsen. 2020.
\newblock \href {http://arxiv.org/abs/2006.10901} {Sparse gpu kernels for deep
  learning}.

\bibitem[{Gao et~al.(2022)Gao, Yu, Li, Dai, Kim, and
  Asanovic}]{gao2022computedram}
Mingyu Gao, Jie Yu, Wentai Li, Michael~C Dai, Nam~Sung Kim, and Krste Asanovic.
  2022.
\newblock computedram: In-memory compute using off-the-shelf dram.
\newblock In \emph{Proceedings of the 27th ACM International Conference on
  Architectural Support for Programming Languages and Operating Systems}, pages
  1065--1079.

\bibitem[{Gemini~Team(2023)}]{team2023gemini}
Google Gemini~Team. 2023.
\newblock Gemini: a family of highly capable multimodal models.
\newblock \emph{arXiv preprint arXiv:2312.11805}.

\bibitem[{Graves(2016)}]{graves2016adaptive}
Alex Graves. 2016.
\newblock Adaptive computation time for recurrent neural networks.
\newblock In \emph{International Conference on Machine Learning}, pages
  3500--3509. PMLR.

\bibitem[{Gunasekar et~al.(2023)Gunasekar, Zhang, Aneja, Mendes, Giorno, Gopi,
  Javaheripi, Kauffmann, de~Rosa, Saarikivi, Salim, Shah, Behl, Wang, Bubeck,
  Eldan, Kalai, Lee, and Li}]{phi_paper}
Suriya Gunasekar, Yi~Zhang, Jyoti Aneja, Caio C{\'{e}}sar~Teodoro Mendes,
  Allie~Del Giorno, Sivakanth Gopi, Mojan Javaheripi, Piero Kauffmann, Gustavo
  de~Rosa, Olli Saarikivi, Adil Salim, Shital Shah, Harkirat~Singh Behl, Xin
  Wang, S{\'{e}}bastien Bubeck, Ronen Eldan, Adam~Tauman Kalai, Yin~Tat Lee,
  and Yuanzhi Li. 2023.
\newblock \href {https://doi.org/10.48550/ARXIV.2306.11644} {Textbooks are all
  you need}.
\newblock \emph{CoRR}, abs/2306.11644.

\bibitem[{Ham et~al.(2016)Ham, Kim, Choi, Cho, Hong, Han, and
  Chung}]{ham2016graphssd}
Jongmin Ham, Jinha Kim, Jinwoong Choi, Cheolwoo Cho, Seulki Hong, Kyeongsu Han,
  and Taejoo Chung. 2016.
\newblock Graphssd: a high performance flash-based storage system for
  large-scale graph processing.
\newblock In \emph{2016 {USENIX} Annual Technical Conference ({USENIX}{ATC}
  16)}, pages 243--256.

\bibitem[{Han et~al.(2016{\natexlab{a}})Han, Liu, Mao, Pu, Pedram, Horowitz,
  and Dally}]{han2016eie}
Song Han, Xingyu Liu, Huizi Mao, Jing Pu, Ardavan Pedram, Mark~A Horowitz, and
  William~J Dally. 2016{\natexlab{a}}.
\newblock Eie: efficient inference engine on compressed deep neural network.
\newblock \emph{arXiv preprint arXiv:1602.01528}.

\bibitem[{Han et~al.(2016{\natexlab{b}})Han, Mao, and Dally}]{han2015deep}
Song Han, Huizi Mao, and William~J Dally. 2016{\natexlab{b}}.
\newblock Deep compression: Compressing deep neural networks with pruning,
  trained quantization and huffman coding.
\newblock In \emph{International Conference on Learning Representations
  (ICLR)}.

\bibitem[{Hannun et~al.(2023)Hannun, Digani, Katharopoulos, and
  Collobert}]{mlx2023}
Awni Hannun, Jagrit Digani, Angelos Katharopoulos, and Ronan Collobert. 2023.
\newblock \href {https://github.com/ml-explore} {{MLX}: Efficient and flexible
  machine learning on apple silicon}.

\bibitem[{He et~al.(2023)He, Zhong, Cai, Lee, and He}]{He2023RESTRS}
Zhenyu He, Zexuan Zhong, Tianle Cai, Jason~D Lee, and Di~He. 2023.
\newblock \href {https://api.semanticscholar.org/CorpusID:265157884} {Rest:
  Retrieval-based speculative decoding}.
\newblock \emph{ArXiv}, abs/2311.08252.

\bibitem[{Hendrycks et~al.(2021)Hendrycks, Burns, Basart, Zou, Mazeika, Song,
  and Steinhardt}]{hendrycks2021measuring}
Dan Hendrycks, Collin Burns, Steven Basart, Andy Zou, Mantas Mazeika, Dawn
  Song, and Jacob Steinhardt. 2021.
\newblock \href {https://openreview.net/forum?id=d7KBjmI3GmQ} {Measuring
  massive multitask language understanding}.
\newblock In \emph{9th International Conference on Learning Representations,
  {ICLR} 2021, Virtual Event, Austria, May 3-7, 2021}. OpenReview.net.

\bibitem[{Hoang et~al.(2023)Hoang, Cho, Merth, Rastegari, and
  Wang}]{Hoang2023DynamicPM}
Duc~Nien Hoang, Minsik Cho, Thomas Merth, Mohammad Rastegari, and Zhangyang
  Wang. 2023.
\newblock \href {https://api.semanticscholar.org/CorpusID:263605807} {(dynamic)
  prompting might be all you need to repair compressed llms}.
\newblock \emph{ArXiv}, abs/2310.00867.

\bibitem[{Holtzman et~al.(2020)Holtzman, Buys, Du, Forbes, and
  Choi}]{Nucleus_Sampling}
Ari Holtzman, Jan Buys, Li~Du, Maxwell Forbes, and Yejin Choi. 2020.
\newblock \href {https://openreview.net/forum?id=rygGQyrFvH} {The curious case
  of neural text degeneration}.
\newblock In \emph{8th International Conference on Learning Representations,
  {ICLR} 2020, Addis Ababa, Ethiopia, April 26-30, 2020}. OpenReview.net.

\bibitem[{Jaiswal et~al.(2023)Jaiswal, Gan, Du, Zhang, Wang, and
  Yang}]{Jaiswal2023CompressingLT}
Ajay Jaiswal, Zhe Gan, Xianzhi Du, Bowen Zhang, Zhangyang Wang, and Yinfei
  Yang. 2023.
\newblock \href {https://api.semanticscholar.org/CorpusID:263605754}
  {Compressing llms: The truth is rarely pure and never simple}.
\newblock \emph{ArXiv}, abs/2310.01382.

\bibitem[{Jiang et~al.(2023)Jiang, Sablayrolles, Mensch, Bamford, Chaplot,
  de~Las~Casas, Bressand, Lengyel, Lample, Saulnier, Lavaud, Lachaux, Stock,
  Scao, Lavril, Wang, Lacroix, and Sayed}]{mistral_paper}
Albert~Q. Jiang, Alexandre Sablayrolles, Arthur Mensch, Chris Bamford,
  Devendra~Singh Chaplot, Diego de~Las~Casas, Florian Bressand, Gianna Lengyel,
  Guillaume Lample, Lucile Saulnier, L{\'{e}}lio~Renard Lavaud, Marie{-}Anne
  Lachaux, Pierre Stock, Teven~Le Scao, Thibaut Lavril, Thomas Wang,
  Timoth{\'{e}}e Lacroix, and William~El Sayed. 2023.
\newblock \href {https://doi.org/10.48550/ARXIV.2310.06825} {Mistral 7b}.
\newblock \emph{CoRR}, abs/2310.06825.

\bibitem[{Leviathan et~al.(2022)Leviathan, Kalman, and
  Matias}]{leviathan2022fast}
Yaniv Leviathan, Matan Kalman, and Yossi Matias. 2022.
\newblock \href {http://arxiv.org/abs/2211.17192} {Fast inference from
  transformers via speculative decoding}.

\bibitem[{Li et~al.(2023)Li, Li, Zhang, and Chu}]{Li2023NormTH}
Liang Li, Qingyuan Li, Bo~Zhang, and Xiangxiang Chu. 2023.
\newblock \href {https://api.semanticscholar.org/CorpusID:261557634} {Norm
  tweaking: High-performance low-bit quantization of large language models}.
\newblock \emph{ArXiv}, abs/2309.02784.

\bibitem[{Lin et~al.(2023)Lin, Tang, Tang, Yang, Dang, and Han}]{Lin2023AWQAW}
Ji~Lin, Jiaming Tang, Haotian Tang, Shang Yang, Xingyu Dang, and Song Han.
  2023.
\newblock \href {https://api.semanticscholar.org/CorpusID:258999941} {Awq:
  Activation-aware weight quantization for llm compression and acceleration}.
\newblock \emph{ArXiv}, abs/2306.00978.

\bibitem[{Liu et~al.(2023{\natexlab{a}})Liu, Oguz, Zhao, Chang, Stock, Mehdad,
  Shi, Krishnamoorthi, and Chandra}]{liu2023llmqat}
Zechun Liu, Barlas Oguz, Changsheng Zhao, Ernie Chang, Pierre Stock, Yashar
  Mehdad, Yangyang Shi, Raghuraman Krishnamoorthi, and Vikas Chandra.
  2023{\natexlab{a}}.
\newblock \href {http://arxiv.org/abs/2305.17888} {Llm-qat: Data-free
  quantization aware training for large language models}.
\newblock \emph{CoRR}.

\bibitem[{Liu et~al.(2023{\natexlab{b}})Liu, Wang, Dao, Zhou, Yuan, Song,
  Shrivastava, Zhang, Tian, Re et~al.}]{liu2023deja}
Zichang Liu, Jue Wang, Tri Dao, Tianyi Zhou, Binhang Yuan, Zhao Song, Anshumali
  Shrivastava, Ce~Zhang, Yuandong Tian, Christopher Re, et~al.
  2023{\natexlab{b}}.
\newblock Deja vu: Contextual sparsity for efficient llms at inference time.
\newblock In \emph{International Conference on Machine Learning}, pages
  22137--22176. PMLR.

\bibitem[{Meswani et~al.(2015)Meswani, Blagodurov, Roberts, Slice, Ignatowski,
  and Loh}]{meswani2015neural}
Moinuddin~K Meswani, Sergey Blagodurov, David Roberts, John Slice, Mike
  Ignatowski, and Gabriel Loh. 2015.
\newblock Neural cache: Bit-serial in-cache acceleration of deep neural
  networks.
\newblock In \emph{2015 48th Annual IEEE/ACM International Symposium on
  Microarchitecture (MICRO)}, pages 383--394. IEEE.

\bibitem[{Mirzadeh et~al.(2023)Mirzadeh, Alizadeh, Mehta, Mundo, Tuzel, Samei,
  Rastegari, and Farajtabar}]{mirzadeh2023relu}
Iman Mirzadeh, Keivan Alizadeh, Sachin Mehta, Carlo C~Del Mundo, Oncel Tuzel,
  Golnoosh Samei, Mohammad Rastegari, and Mehrdad Farajtabar. 2023.
\newblock \href {http://arxiv.org/abs/2310.04564} {Relu strikes back:
  Exploiting activation sparsity in large language models}.

\bibitem[{Parashar et~al.(2017)Parashar, Rhu, Mukkara, Puglielli, Venkatesan,
  Khailany, Emer, Keckler, and Dally}]{parashar2017timeloop}
Angshuman Parashar, Minsoo Rhu, Anurag Mukkara, Antonio Puglielli, Rangharajan
  Venkatesan, Brucek Khailany, Joel Emer, Stephen~W Keckler, and William~J
  Dally. 2017.
\newblock Timeloop: A systematic approach to dnn accelerator evaluation.
\newblock In \emph{2017 IEEE International Symposium on Performance Analysis of
  Systems and Software (ISPASS)}, pages 241--251. IEEE.

\bibitem[{Paszke et~al.(2019)Paszke, Gross, Massa, Lerer, Bradbury, Chanan,
  Killeen, Lin, Gimelshein, Antiga, Desmaison, K{\"{o}}pf, Yang, DeVito,
  Raison, Tejani, Chilamkurthy, Steiner, Fang, Bai, and Chintala}]{pytorch}
Adam Paszke, Sam Gross, Francisco Massa, Adam Lerer, James Bradbury, Gregory
  Chanan, Trevor Killeen, Zeming Lin, Natalia Gimelshein, Luca Antiga, Alban
  Desmaison, Andreas K{\"{o}}pf, Edward~Z. Yang, Zachary DeVito, Martin Raison,
  Alykhan Tejani, Sasank Chilamkurthy, Benoit Steiner, Lu~Fang, Junjie Bai, and
  Soumith Chintala. 2019.
\newblock \href
  {https://proceedings.neurips.cc/paper/2019/hash/bdbca288fee7f92f2bfa9f7012727740-Abstract.html}
  {Pytorch: An imperative style, high-performance deep learning library}.
\newblock In \emph{Advances in Neural Information Processing Systems 32: Annual
  Conference on Neural Information Processing Systems 2019, NeurIPS 2019,
  December 8-14, 2019, Vancouver, BC, Canada}, pages 8024--8035.

\bibitem[{Rajbhandari et~al.(2021)Rajbhandari, Ruwase, Rasley, Smith, and
  He}]{rajbhandari2021zero}
Samyam Rajbhandari, Olatunji Ruwase, Jeff Rasley, Shaden Smith, and Yuxiong He.
  2021.
\newblock Zero-infinity: Breaking the gpu memory wall for extreme scale deep
  learning.
\newblock In \emph{SC21: International Conference for High Performance
  Computing, Networking, Storage and Analysis}, pages 1--14.

\bibitem[{Rhu et~al.(2013)Rhu, Gimelshein, Clemons, Zulfiqar, and
  Keckler}]{rhu2013vdnn}
Minsoo Rhu, Natalia Gimelshein, Jason Clemons, Arslan Zulfiqar, and Stephen~W
  Keckler. 2013.
\newblock vdnn: Virtualized deep neural networks for scalable, memory-efficient
  neural network design.
\newblock In \emph{2016 49th Annual IEEE/ACM International Symposium on
  Microarchitecture (MICRO)}, page Article 13. IEEE Computer Society.

\bibitem[{Shao et~al.(2023)Shao, Chen, Zhang, Xu, Zhao, Li, Zhang, Gao, Qiao,
  and Luo}]{Shao2023OmniQuantOC}
Wenqi Shao, Mengzhao Chen, Zhaoyang Zhang, Peng Xu, Lirui Zhao, Zhiqiang Li,
  Kaipeng Zhang, Peng Gao, Yu~Jiao Qiao, and Ping Luo. 2023.
\newblock \href {https://api.semanticscholar.org/CorpusID:261214575}
  {Omniquant: Omnidirectionally calibrated quantization for large language
  models}.
\newblock \emph{ArXiv}, abs/2308.13137.

\bibitem[{Shao et~al.(2022)Shao, Li, Cai, Wang, Narayanan, and
  Ranganathan}]{shao2022hotpot}
Yifan Shao, Mengjiao Li, Wenhao Cai, Qi~Wang, Dhananjay Narayanan, and
  Parthasarathy Ranganathan. 2022.
\newblock Hotpot: Warmed-up gigascale inference with tightly-coupled compute
  and reuse in flash.
\newblock In \emph{Proceedings of the 55th Annual IEEE/ACM International
  Symposium on Microarchitecture}, pages 335--349.

\bibitem[{Sheng et~al.(2023)Sheng, Zheng, Yuan, Li, Ryabinin, Chen, Liang,
  R{\'{e}}, Stoica, and Zhang}]{sheng2023flexgen}
Ying Sheng, Lianmin Zheng, Binhang Yuan, Zhuohan Li, Max Ryabinin, Beidi Chen,
  Percy Liang, Christopher R{\'{e}}, Ion Stoica, and Ce~Zhang. 2023.
\newblock Flexgen: High-throughput generative inference of large language
  models with a single {GPU}.
\newblock In \emph{International Conference on Machine Learning, {ICML} 2023,
  23-29 July 2023, Honolulu, Hawaii, {USA}}, volume 202 of \emph{Proceedings of
  Machine Learning Research}, pages 31094--31116. {PMLR}.

\bibitem[{Song et~al.(2024)Song, Han, Zhang, Hu, Shi, Li, Chen, Liu, Li, Yang,
  and Sun}]{song2024prosparse}
Chenyang Song, Xu~Han, Zhengyan Zhang, Shengding Hu, Xiyu Shi, Kuai Li, Chen
  Chen, Zhiyuan Liu, Guangli Li, Tao Yang, and Maosong Sun. 2024.
\newblock \href {http://arxiv.org/abs/2402.13516} {Prosparse: Introducing and
  enhancing intrinsic activation sparsity within large language models}.

\bibitem[{Subramani et~al.(2022)Subramani, Savvides, Ping, and
  Narang}]{subramani2022adapt}
Vedant Subramani, Marios Savvides, Li~Ping, and Sharan Narang. 2022.
\newblock Adapt: Parameter adaptive token-wise inference for vision
  transformers.
\newblock In \emph{Proceedings of the 55th Annual IEEE/ACM International
  Symposium on Microarchitecture}.

\bibitem[{Sun et~al.(2023)Sun, Liu, Bair, and Kolter}]{Sun2023ASA}
Mingjie Sun, Zhuang Liu, Anna Bair, and J.~Zico Kolter. 2023.
\newblock \href {https://api.semanticscholar.org/CorpusID:259203115} {A simple
  and effective pruning approach for large language models}.
\newblock \emph{ArXiv}, abs/2306.11695.

\bibitem[{Touvron et~al.(2023{\natexlab{a}})Touvron, Lavril, Izacard, Martinet,
  Lachaux, Lacroix, Rozi{\`{e}}re, Goyal, Hambro, Azhar, Rodriguez, Joulin,
  Grave, and Lample}]{Llamav1paper}
Hugo Touvron, Thibaut Lavril, Gautier Izacard, Xavier Martinet, Marie{-}Anne
  Lachaux, Timoth{\'{e}}e Lacroix, Baptiste Rozi{\`{e}}re, Naman Goyal, Eric
  Hambro, Faisal Azhar, Aur{\'{e}}lien Rodriguez, Armand Joulin, Edouard Grave,
  and Guillaume Lample. 2023{\natexlab{a}}.
\newblock \href {http://arxiv.org/abs/2302.13971} {Llama: Open and efficient
  foundation language models}.
\newblock \emph{CoRR}, abs/2302.13971.

\bibitem[{Touvron et~al.(2023{\natexlab{b}})Touvron, Martin, Stone, Albert,
  Almahairi, Babaei, Bashlykov, Batra, Bhargava, Bhosale
  et~al.}]{touvron2023llama}
Hugo Touvron, Louis Martin, Kevin Stone, Peter Albert, Amjad Almahairi, Yasmine
  Babaei, Nikolay Bashlykov, Soumya Batra, Prajjwal Bhargava, Shruti Bhosale,
  et~al. 2023{\natexlab{b}}.
\newblock Llama 2: Open foundation and fine-tuned chat models.
\newblock \emph{arXiv preprint arXiv:2307.09288}.

\bibitem[{Wolf et~al.(2019)Wolf, Debut, Sanh, Chaumond, Delangue, Moi, Cistac,
  Rault, Louf, Funtowicz, and Brew}]{huggingface_transformers}
Thomas Wolf, Lysandre Debut, Victor Sanh, Julien Chaumond, Clement Delangue,
  Anthony Moi, Pierric Cistac, Tim Rault, R{\'{e}}mi Louf, Morgan Funtowicz,
  and Jamie Brew. 2019.
\newblock \href {http://arxiv.org/abs/1910.03771} {Huggingface's transformers:
  State-of-the-art natural language processing}.
\newblock \emph{CoRR}, abs/1910.03771.

\bibitem[{Xia et~al.(2023)Xia, Zheng, Li, Zhuang, Zhou, Qiu, Li, Lin, and
  Song}]{Xia2023FlashLLMEL}
Haojun Xia, Zhen Zheng, Yuchao Li, Donglin Zhuang, Zhongzhu Zhou, Xiafei Qiu,
  Yong Li, Wei Lin, and Shuaiwen~Leon Song. 2023.
\newblock \href {https://api.semanticscholar.org/CorpusID:262054394}
  {Flash-llm: Enabling low-cost and highly-efficient large generative model
  inference with unstructured sparsity}.
\newblock \emph{Proc. VLDB Endow.}, 17:211--224.

\bibitem[{Xu et~al.(2023)Xu, Liu, Chen, Tang, Wang, Zhou, Hu, and
  Shrivastava}]{Xu2023CompressTP}
Zhaozhuo Xu, Zirui Liu, Beidi Chen, Yuxin Tang, Jue Wang, Kaixiong Zhou, Xia
  Hu, and Anshumali Shrivastava. 2023.
\newblock \href {https://api.semanticscholar.org/CorpusID:258823240} {Compress,
  then prompt: Improving accuracy-efficiency trade-off of llm inference with
  transferable prompt}.
\newblock \emph{ArXiv}, abs/2305.11186.

\bibitem[{Yi et~al.(2023)Yi, Guo, Wei, Zhou, Wang, and Xu}]{Yi2023EdgeMoEFO}
Rongjie Yi, Liwei Guo, Shiyun Wei, Ao~Zhou, Shangguang Wang, and Mengwei Xu.
  2023.
\newblock \href {https://api.semanticscholar.org/CorpusID:261243273} {Edgemoe:
  Fast on-device inference of moe-based large language models}.
\newblock \emph{ArXiv}, abs/2308.14352.

\bibitem[{Zhang et~al.(2023{\natexlab{a}})Zhang, Ning, Prabhakar, and
  Wentzlaff}]{Zhang2023AHE}
Hengrui Zhang, August Ning, Rohan Prabhakar, and David Wentzlaff.
  2023{\natexlab{a}}.
\newblock \href {https://api.semanticscholar.org/CorpusID:265711466} {A
  hardware evaluation framework for large language model inference}.

\bibitem[{Zhang et~al.(2023{\natexlab{b}})Zhang, Wang, Li, Shou, Chen, Chen,
  and Mehrotra}]{Zhang2023DraftV}
Jinchao Zhang, Jue Wang, Huan Li, Lidan Shou, Ke~Chen, Gang Chen, and Sharad
  Mehrotra. 2023{\natexlab{b}}.
\newblock \href {https://api.semanticscholar.org/CorpusID:262013673} {Draft \&
  verify: Lossless large language model acceleration via self-speculative
  decoding}.
\newblock \emph{ArXiv}, abs/2309.08168.

\bibitem[{Zhang et~al.(2022{\natexlab{a}})Zhang, Dai, Sheng, Zhang, Li, Xu,
  Dai, Xiao, Ma, Tang et~al.}]{zhang2022llm}
Shizhao Zhang, Han Dai, Tian Sheng, Jiawei Zhang, Xiaoyong Li, Qun Xu, Mengjia
  Dai, Yunsong Xiao, Chao Ma, Rui Tang, et~al. 2022{\natexlab{a}}.
\newblock Llm quantization: Quantization-aware training for large language
  models.
\newblock In \emph{Advances in Neural Information Processing Systems},
  volume~35.

\bibitem[{Zhang et~al.(2022{\natexlab{b}})Zhang, Roller, Goyal, Artetxe, Chen,
  Chen, Dewan, Diab, Li, Lin, Mihaylov, Ott, Shleifer, Shuster, Simig, Koura,
  Sridhar, Wang, and Zettlemoyer}]{OPTpaper}
Susan Zhang, Stephen Roller, Naman Goyal, Mikel Artetxe, Moya Chen, Shuohui
  Chen, Christopher Dewan, Mona~T. Diab, Xian Li, Xi~Victoria Lin, Todor
  Mihaylov, Myle Ott, Sam Shleifer, Kurt Shuster, Daniel Simig, Punit~Singh
  Koura, Anjali Sridhar, Tianlu Wang, and Luke Zettlemoyer. 2022{\natexlab{b}}.
\newblock \href {http://arxiv.org/abs/2205.01068} {{OPT:} open pre-trained
  transformer language models}.
\newblock \emph{CoRR}, abs/2205.01068.

\bibitem[{Zhang et~al.(2024)Zhang, Song, Yu, Han, Lin, Xiao, Song, Liu, Mi, and
  Sun}]{zhang2024relu2}
Zhengyan Zhang, Yixin Song, Guanghui Yu, Xu~Han, Yankai Lin, Chaojun Xiao,
  Chenyang Song, Zhiyuan Liu, Zeyu Mi, and Maosong Sun. 2024.
\newblock \href {http://arxiv.org/abs/2402.03804} {Relu$^2$ wins: Discovering
  efficient activation functions for sparse llms}.

\bibitem[{Zhao et~al.(2023)Zhao, Lin, Zhu, Ye, Chen, Zheng, Ceze,
  Krishnamurthy, Chen, and Kasikci}]{Zhao2023AtomLQ}
Yilong Zhao, Chien-Yu Lin, Kan Zhu, Zihao Ye, Lequn Chen, Size Zheng, Luis
  Ceze, Arvind Krishnamurthy, Tianqi Chen, and Baris Kasikci. 2023.
\newblock \href {https://api.semanticscholar.org/CorpusID:264828796} {Atom:
  Low-bit quantization for efficient and accurate llm serving}.
\newblock \emph{ArXiv}, abs/2310.19102.

\end{thebibliography}

\clearpage
\appendix

\section{Appendix Overview}
The appendix is structured as follows:
\begin{itemize}[leftmargin=*]

\item In Appendix~\ref{sec:appendix-predictor}, we provide additional details on the low-rank predictor introduced in Section~\ref{sec:load_from_flash}. We evaluate our trained predictors from both accuracy (i.e., their impact on the model's accuracy) and efficiency perspectives (i.e., the additional neurons they predict to be activated).

\item Appendix~\ref{sec:appendix_results} offers a more detailed description of our experimental setup and implementation for the experiments conducted in Section~\ref{sec:experiments-and-results}.
\item In Appendix~\ref{sec:appendix-bundling-coactivation}, we discuss a negative result regarding the strategy of bundling neurons based on co-activation as a potential method for increasing chunk size (cf. Section~\ref{sec:increase-chunk-size}). We intentionally include this negative result as we believe it may inspire future research on effective neuron bundling and its utilization for efficient inference.
\item In Appendix~\ref{sec:appendix-related-works}, we delve deeper into the review of related works in the literature.
\item In Appendix~\ref{sec:small-device}, we go over implications of llm in flash when going to smaller devices.
\item Finally, Appendix~\ref{sec:qualitative-evals} compares the texts generated by the base model with those produced by our models that utilize the predictor.
\end{itemize}

\section{Low-Rank Activation Predictor: Additional Results}
\label{sec:appendix-predictor}

\begin{figure*}[t]
\centering
\begin{subfigure}{.245\textwidth}
  \centering
  \includegraphics[width=\textwidth]{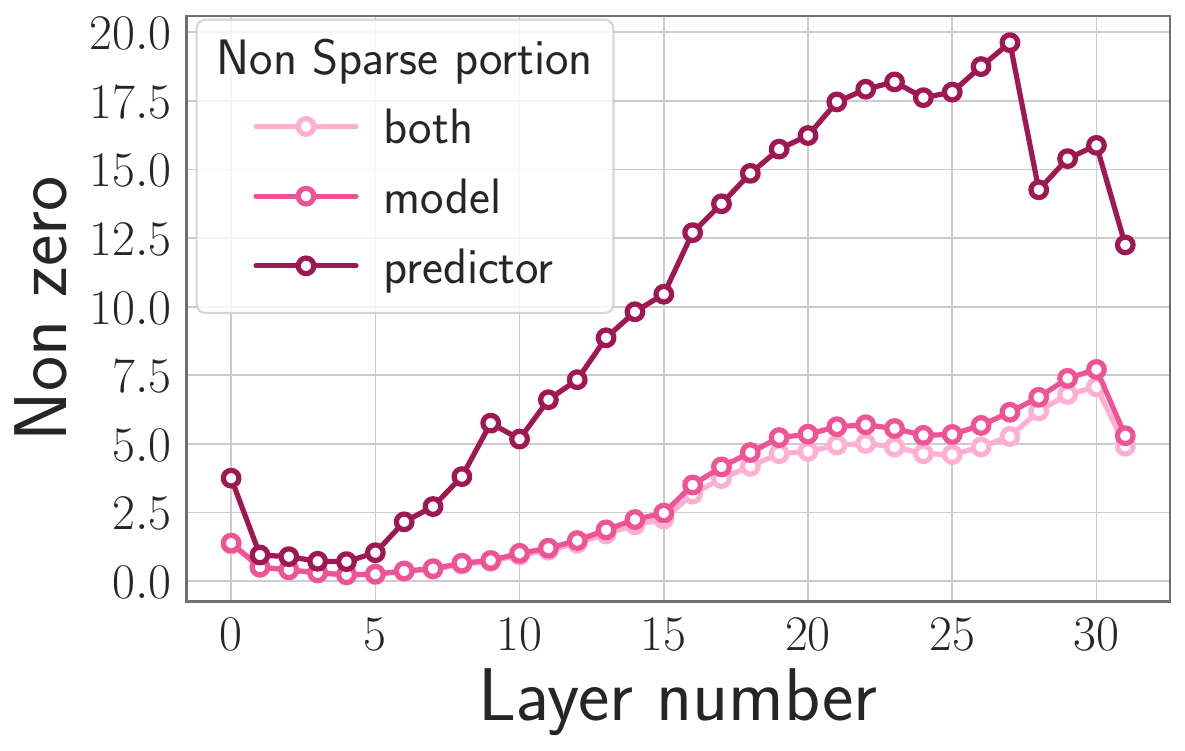}
\caption{OPT-6.7B}
  \label{fig:sparsity_models_opt6.7b}
\end{subfigure}\hfill
\begin{subfigure}{.245\textwidth}
  \centering
  \includegraphics[width=\textwidth]{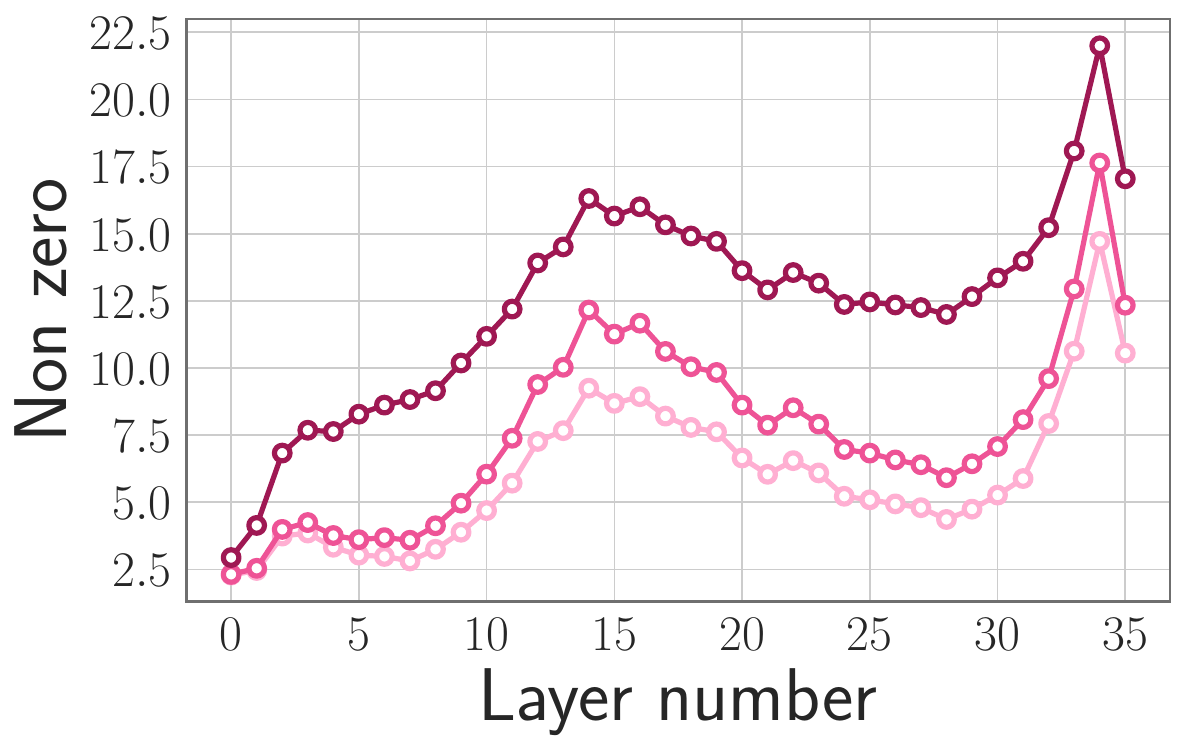}
    \caption{Persimmon 8B}
  \label{fig:sparsity_models_persimmon}
\end{subfigure}\hfill
\begin{subfigure}{.245\textwidth}
  \centering
  \includegraphics[width=\textwidth]{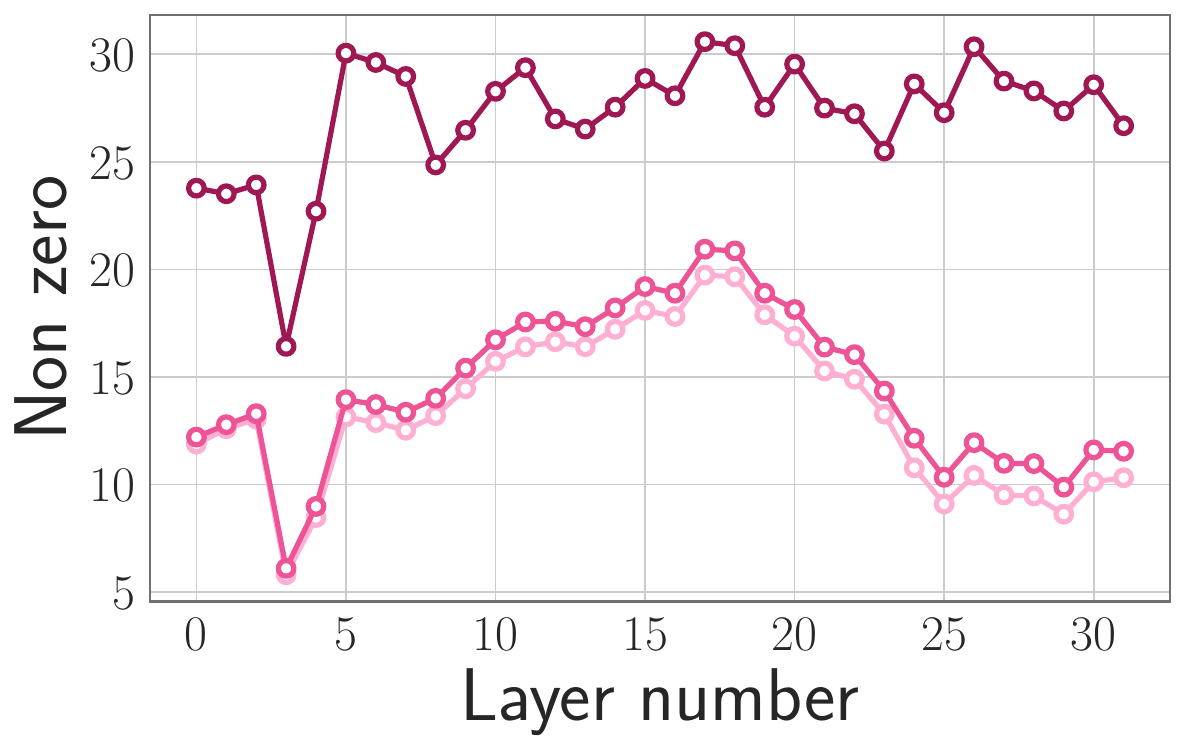}
    \caption{Phi-2}
  \label{fig:sparsity_models_phi2}
\end{subfigure}\hfill
\begin{subfigure}{.245\textwidth}
  \centering
  \includegraphics[width=\textwidth]{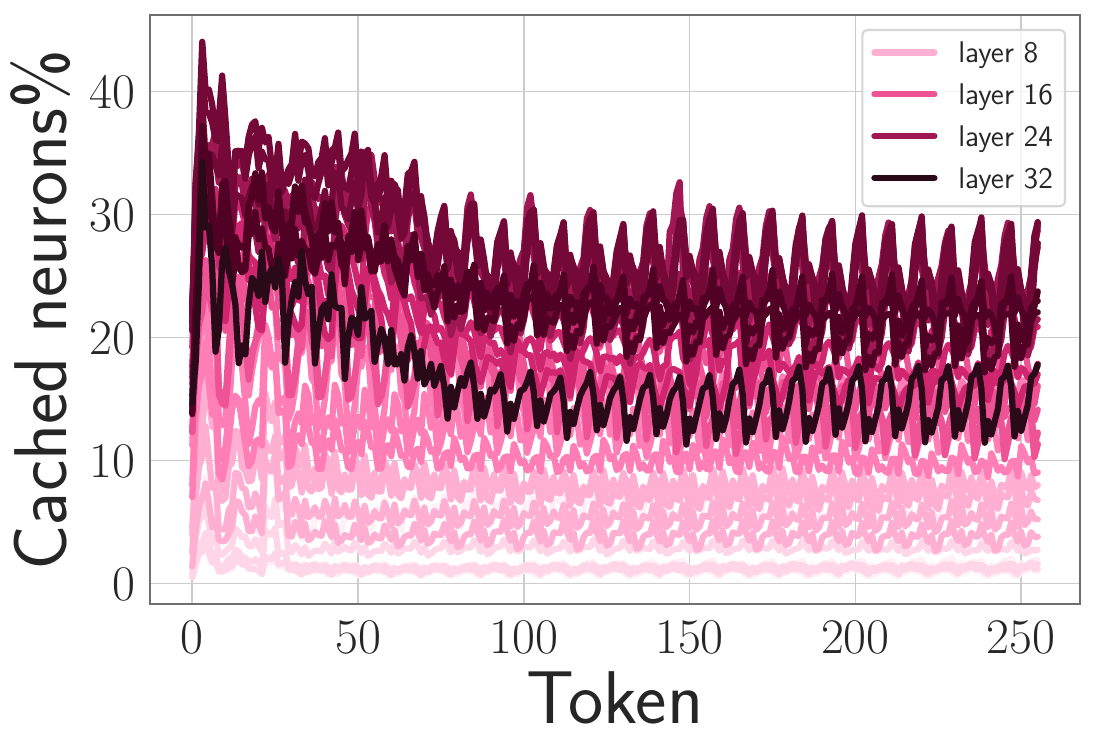}
    \caption{Rows cached over time}
  \label{fig:sparsity_models_rows_cached}
\end{subfigure}\hfill
\caption{\textbf{(a)} 
The percentage of fired neurons in each layer's FFN is less than 5\%. In predictor, roughly 3x of this amount will get activated. The narrow gap between neurons that are activated in both shows the model output will not change abruptly.  \textbf{(b)} Earlier layers of Persimmon have less active neurons, and later layers of Persimmon have higher active neurons, so we trained larger predictors for them. \textbf{(c)} In Phi2 middle layers have a higher active neuron ratio, so we trained larger predictors for those layers. \textbf{(d)} The number of neurons cached within a real scenario of inference, later layers have more cached rows because of their higher nonsparse ratio.}
\label{fig:sparsity_models}
\end{figure*}

\subsection{Sparsity patterns of predictors}
The number of neurons predicted to be active will determine the efficiency of our algorithm, the less sparse the predicted activation the more weights will have to be loaded from flash. We evaluated the sparsity patterns over 100 random samples of the C4 validation dataset. In \figref{fig:sparsity_models} we can see the sparsity patterns of OPT, Persimmon, and Phi. In OPT, the number of active neurons predicted by the predictor is 3x the amount of actual sparsity observed in the case of dense inference. In Persimmon it is about the same - 3x the required neurons, and in Phi-2 it is roughly 2x the required neurons of the original model that are activated by the predictor. The neurons that are activated by the model and not the predictor are the false negatives. The gap between the neurons active in both the predictor and the model, and the neurons active only in the model is very narrow in all three models, hence false negatives constitute a small fraction of predictions. To reduce false negatives, the predictor has to "over-predict", which results in loading neurons that are redundant, that is, will have zero activation and no effect on the outcome. An interesting future direction of this work is improving the accuracy of the predictors to be able to load fewer neurons. One observation we had in OPT and Persimmon is the later layers have more active neurons, which can be seen in Figure \ref{fig:sparsity_models_rows_cached}.

\subsection{Accuracy of models using predictors}
\begin{figure*}[t]
\centering
\begin{subfigure}{.42\textwidth}
  \centering
  \includegraphics[width=\textwidth]{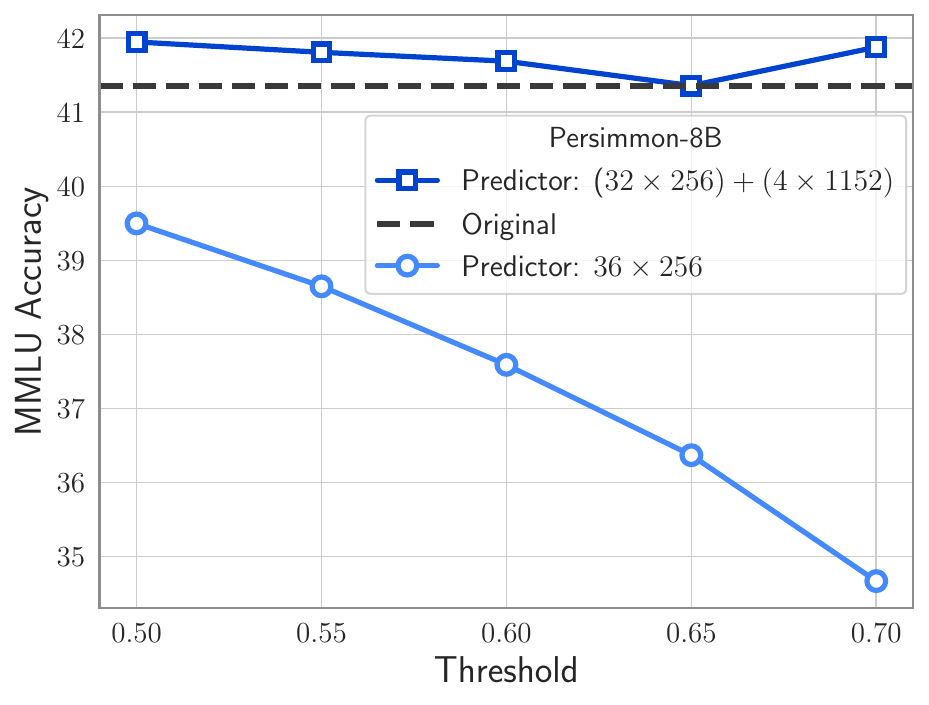}
\caption{}
  \label{fig:mmlu_persimmon}
\end{subfigure}\hfill
\begin{subfigure}{.43\textwidth}
  \centering
  \includegraphics[width=\textwidth]{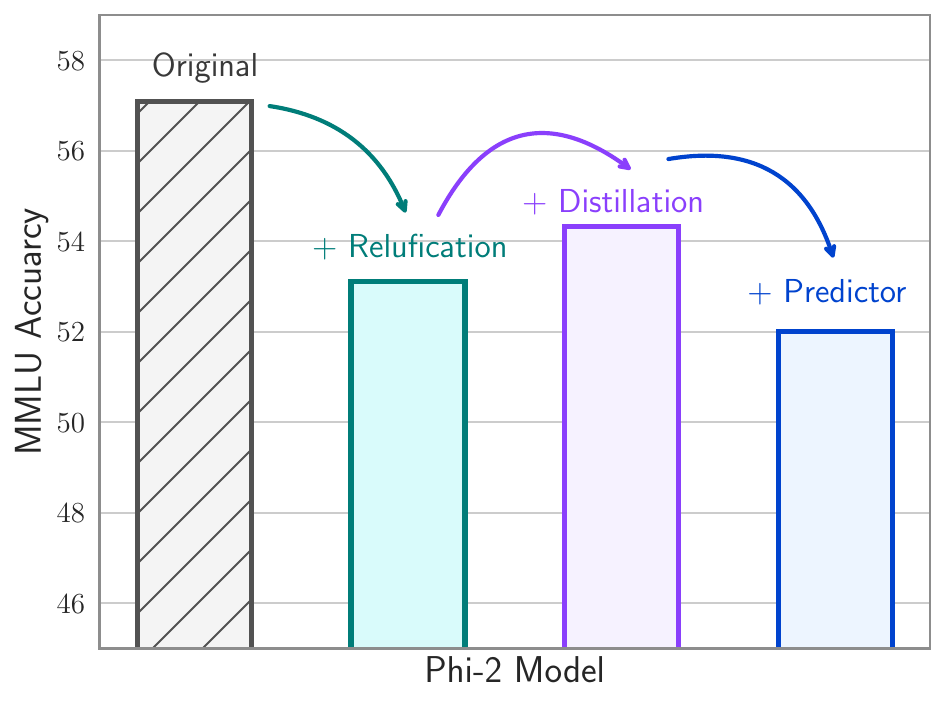}
    \caption{}
  \label{fig:mmlu_phi}
\end{subfigure}\hfill

\caption{\textbf{(a)} If we use larger predictors in the last 4 layers MMLU wouldn't drop a lot in Persimmon. \textbf{(b)} Phi's MMLU will drop in the relufication process due to lower quality data, using distillation can improve the results for that. Using predictors will downgrade the results but still keep it at 52.}
\label{fig:mmlu}
\end{figure*}

We evaluate the accuracy of models on public benchmarks with predictors in place. In \tabref{tab:model_performance_zeroshot} it can be seen zero shot accuracy of models doesn't drop. Also, we can see that increasing the predictor size for the last 4 layers of Persimmon and Falcon improves the zero-shot metrics. We evaluated models on MMLU \citep{hendrycks2021measuring} benchmark as well. We used Instruct Eval's implementaion \citep{chia2023instructeval} for evaluating MMLU. In \figref{fig:mmlu_persimmon} we can see the MMLU of Persimmon doesn't drop when the last 4 layers use higher rank predictors but this is not the case for lower ranked ones. Phi2's MMLU will drop 2.3 points from the relufied model still keeping at 52 as shown in \figref{fig:mmlu_phi}. By increasing the threshold of low-rank predictor we can reduce the amount of data load, this comes with a slight degradation in zero-shot metrics as seen in the \tabref{tab:model_performance_zeroshot} for different thresholds of the Persimmon model. We have used threshold=0.7 for Persimmon.

\subsection{Overhead of predictors}
The average rank of predictors in the OPT-6.7B is 240, this will result in less than 2.4\% of non-embedding weights and FLOPs. In M1 Max CPU experiments this was comprising 2.75\% and in RTX GPU it was 4.8\% of inference time which is negligible. For Falocn 7B, predictors take ~4\% model size and CPU computation. For Persimmon it was taking 2.85\% of inference time on CPU. For Llama 2 7B it was taking 3.92\% of inference time on CPU.

\begin{table*}[htbp]
\centering
\caption{Model performance on zero-shot tasks when using predictors}
\label{tab:model_performance_zeroshot}
\begin{threeparttable}
\resizebox{1.0\textwidth}{!}{%
\begin{tabular}{@{}l|l l l|l l l@{}}
\toprule
\textbf{Model} & \multicolumn{3}{c|}{\textbf{Predictor Parameters}} & \multicolumn{3}{c}{\textbf{Zero-Shot Metrics}} \\
\midrule
& \textbf{\begin{tabular}[c]{@{}l@{}}Rank for\\Sensitives \tnote{*}\end{tabular}} & \textbf{\begin{tabular}[c]{@{}l@{}}Rank for\\Other Layers\end{tabular}} & \textbf{Threshold} & \textbf{ArcEasy} & \textbf{\begin{tabular}[c]{@{}l@{}}Arc\\Challenge\end{tabular}} & \textbf{\begin{tabular}[c]{@{}l@{}}Hella\\Swag\end{tabular}} \\
\midrule
\textbf{OPT 6.7B} & - & - & - & 66.1 & 30.6 & 50.3 \\
\textbf{OPT 6.7B with predictors} & 1024 & 128 & 0.5 & 66.2 & 30.6 & 49.8 \\
\midrule
\textbf{Falcon 7B} & - & - & - & 74.62 & 40.05 & 57.77 \\
\textbf{Falcon 7B relufied} & - & - & - & 72.52 & 38.23 & 54.17 \\
\textbf{Falcon 7B relufied with predictors} & 128 & 128 & 0.50 & 70.20 & 35.41 & 50.74 \\
\textbf{Falcon 7B relufied with predictors} & 1152 & 128 & 0.50 & 71.51 & 34.22 & 52.28 \\
\textbf{Falcon 7B relufied with predictors} & 1152 & 256 & 0.50 & 72.35 & 36.35 & 53.16 \\
\midrule
\textbf{Persimmon 8B} & - & - & - & 67.80 & 34.64 & 50.70 \\
\textbf{Persimmon 8B with predictors} & 256 & 256 & 0.5 & 67.26 & 33.87 & 50.51 \\
\textbf{Persimmon 8B with predictors} & 256 & 256 & 0.55 & 66.71 & 34.73 & 50.54 \\
\textbf{Persimmon 8B with predictors} & 256 & 256 & 0.60 & 66.67 & 34.04 & 50.59 \\
\textbf{Persimmon 8B with predictors} & 256 & 256 & 0.65 & 66.41 & 34.22 & 50.42 \\
\textbf{Persimmon 8B with predictors} & 1152 & 256 & 0.70 & 66.30 & 34.40 & 52.70 \\
\midrule
\textbf{Phi-2} & -  & - & - & 79.62 & 51.49 & 55.17 \\
\textbf{Phi-2 relufied} & - & - & - & 80.60 & 50.12 & 54.30 \\
\textbf{Phi-2 relufied with predictors} & 800 & mix 160, 480 & 0.40 & 79.96 & 49.57 & 53.50 \\
\textbf{Phi-2 relufied with predictors} & 800 & mix 160, 480  & 0.55 & 78.90 & 47.90 & 52.75 \\
\bottomrule
\end{tabular}%
}
\begin{tablenotes}
    \item[*] For OPT, Falcon, and Persimmon sensitive layers are the last 4 layers. For Phi-2 it is the middle 8.
\end{tablenotes}
\end{threeparttable}

\end{table*}

\section{Extended Results}
\label{sec:appendix_results}

\textbf{Experimental Setup:}
Our experiment is designed to optimize inference efficiency on personal devices. To this end, we process sequences individually, running only one sequence at a time. This approach allows us to allocate a specific portion of DRAM for the Key-Value (KV) cache while primarily focusing on the model size. This strategy is particularly effective when dealing with only one sequence/query at a time.\footnote{ For OPT 6.7 B model with context length 2048 KV-cache requires $2048\times 2d_{\text{model}}$ elements which is only $8\%$ of model size. Also, the KV cache itself can be held in flash memory.}

For the implementation of our inference process, we utilize the HuggingFace Transformers and KV caching. This setup is tested under the condition that approximately half of the model size is available in DRAM. We select this amount as a showcase of the idea of hosting the LLM in Flash. With a different level of sparsity or employing quantization, one can work with smaller available DRAM capacity as well, or alternatively use larger models. Such a configuration demonstrates the practicality of executing inference with lower memory footprints.

\textbf{Systems performance optimization:}
The primary target of our experiments was the Apple macOS 14.3 operating system. For high-performance inference, most of the existing deep learning frameworks require that the shape of the weights and the intermediate results in the computation remain static throughout. In particular, the Metal Performance Shaders (MPS) backend for PyTorch demonstrates rather steep performance cliffs when any shape dynamism is present in the computational graph. In order to build a high-performance implementation, we chose to borrow custom, dynamism-friendly Metal kernels from Apple's open-source MLX deep learning framework~\citep{mlx2023}. In addition, we made use of the unified memory architecture available on Apple systems, which we exploit to maintain the weights cache using the GPU allocator, by creating the tensors using the \texttt{MTLStorageModeShared} allocation mode. This mode allows both the CPU and the GPU to access the same memory buffer directly, without redundant copies. We observe that the inputs to and the outputs from the feed-forward network have a static shape, so by hiding the dynamism inside a binary extension and handling the shape dynamism and memory management internally, we were able to achieve a level of performance that is not achievable with PyTorch MPS backend alone while leaving the rest of the model intact. Over the course of our work, we were able to eliminate nearly all redundant data movement, improving inference performance.

\begin{table*}[t]
\centering
\caption{The end-to-end inference latency across different setups with standard deviation. Our efficient implementation (referred to as \emph{All}) that employs the predictor, windowing, and bundling can lead to significant latency reduction.}
\label{tab:full-e2e-results-with-std}
\resizebox{1\textwidth}{!}{%
\begin{tabular}{@{}lll|cccc@{}}
\cmidrule(l){4-7}
                                    &    &     & \multicolumn{4}{c}{\textbf{Inference Latency (ms)}}                                            \\ \midrule
\multicolumn{1}{l|}{\textbf{Model}} & \textbf{Method} & \textbf{Backend} & \textbf{I/O} & \textbf{Mem} & \textbf{Compute} & \textbf{Total} \\ \midrule
\multicolumn{1}{l|}{OPT 6.7B}       & All & CPU       & \textbf{104.90 ($\pm$ 18.46)}                      & 57.79 ($\pm$ 9.63)                         & 506.50 ($\pm$17.33)              & \textbf{669.20 ($\pm$ 39.74) }            \\ \hdashline

\multicolumn{1}{l|}{OPT 6.7B}       & All & GPU        & \textbf{30.55 ($\pm$3.09)   }                       & 34.11 ($\pm$2.38)                            & 19.97 ($\pm$0.86)              & 84.64 ($\pm$6.16)                \\ 
\multicolumn{1}{l|}{OPT 6.7B}       & Speculative & GPU        & 38.53 ($\pm$10.0)                       & 9.45 ($\pm$1.7)                           & 12.18 ($\pm$2.0)                  & \textbf{60.16 ($\pm$13.4)   }             \\ \midrule

\multicolumn{1}{l|}{Persimmon 8B}   & All & CPU        & \textbf{310.52 ($\pm$41.12)}                      & 155.80 ($\pm$21.30)                         & 623.74 ($\pm$ 24.76)            & \textbf{1090.08 ($\pm$79.08)}           \\ \midrule

\multicolumn{1}{l|}{Phi-2}          & All & CPU       & \textbf{211.08 ($\pm$24.81)}                      & 76.87  ($\pm$7.18)                        & 258.74 ($\pm20.90$)              & \textbf{546.69 ($\pm$31.98)}            \\ \bottomrule
\end{tabular}%
}
\end{table*}

\textbf{Caching Considerations for Data Loading from Flash Memory.}
When data is read from flash memory, the operating system typically caches the blocks in the block cache, anticipating future reuse. However, this caching mechanism consumes additional memory in DRAM beyond what is allocated for the model. To accurately assess the real throughput of flash memory under limited DRAM conditions, benchmarks should be conducted without relying on caching. Practical systems may or may not rely on filesystem cache, depending on requirements.

\looseness=-1 For the purpose of our hardware benchmarking in this study, we deliberately and significantly pessimize our NVMe throughput measurements. On macOS and iOS, we employ the \texttt{F\_NOCACHE} flag with the \texttt{fcntl()} function, while on Linux, we use  \texttt{DirectIO}. Additionally, on macOS, we clear any resident buffers before initiating the benchmark using the \texttt{purge} command. This approach provides a conservative lower bound of throughput in scenarios where no caching is permitted and makes the benchmarks repeatable. It's worth noting that these figures can improve if either the inference code or the operating system is allowed to cache some part of the weights.

While OS-level buffer caching is advantageous for general-purpose applications with high cache hit rates, it lacks fine-grained control over cache usage per process or buffer eviction at the application level. In the context of on-device memory constraints and large model sizes, this could lead to a situation where the file system level cache does not help because in order to evaluate later layers earlier layers must be evicted in a rolling pattern, so the effective cache hit rate is close to zero. Aside from being inefficient, this can cause coexistence issues with other processes due to memory allocation pressure and Translation Lookaside Buffer (TLB) churn.

\subsection{Results for OPT 6.7B Model}

This section presents the outcomes for the OPT 6.7B model, specifically under conditions where the memory allocated for the model in DRAM is approximately half of its baseline requirement. 

\textbf{Predictors.}
For the initial 28 layers of the OPT 6.7B model, we train predictors with a rank of $r=128$. To reduce the occurrence of false negatives, the final four layers employ predictors with a higher rank of $r=1024$. These predictors achieve an average of 5\% false negatives and 7\% false positives in the OPT 6.7B model. As depicted in Figure \ref{fig:act_compare_sparsity}, our predictor accurately identifies most activated neurons, while occasionally misidentifying inactive ones with values near zero. Notably, these false negatives, being close to zero, do not significantly alter the final output when they are excluded. Furthermore, as demonstrated in \tabref{tab:accuracy_metrics}, this level of prediction accuracy does not adversely affect the model's performance in 0-shot tasks.

\textbf{Windowing in the OPT 6.7B Model.}
Utilizing a windowing method with \( k=4 \) in the OPT 6.7B model significantly reduces the necessity for fresh data loading. Using active neurons of predictor would require about 10\% of the DRAM memory capacity on average; however, with our method, it drops to 2.4\%. This process involves reserving DRAM memory for a window of the past 5 tokens, which, in turn, increases the DRAM requirement for the Feed Forward Network (FFN) to 24\%.

The overall memory retained in DRAM for the model comprises several components: Embeddings, the Attention Model, the Predictor, and the Loaded Feed Forward layer. The Predictor accounts for 1.25\% of the model size, while Embeddings constitute 3\%. The Attention Model's weights make up 32.3\%, and the FFN occupies 15.5\% (calculated as \(0.24 \times 64.62\)). Summing these up, the total DRAM memory usage amounts to 52.1\% of the model's size.

\textbf{Latency Analysis:}
Using a window size of 4, each token requires access to 2.4\% of the Feed Forward Network (FFN) neurons. For a 32-bit model, the data chunk size per read is \(2d_{\text{model}} \times 4 \text{ bytes} = 32 \text{ KiB}\), as it involves concatenated rows and columns. On an M1 Max, this results in the average latency of \(105ms\) per token for loading from flash and \(57ms\) for memory management (involving neuron deletion and addition). Thus, the total memory-related latency is less than \(162ms\) per token (refer to Figure \ref{fig:latency-intro}). In contrast, the baseline approach, which requires loading 13.4GB of data at a speed of 6.1GB/s, leads to a latency of approximately \(2196ms\) per token. Therefore, our method represents a substantial improvement over the baseline.

For a 16-bit model on a GPU machine, the flash load time is reduced to \(30.5ms\), and memory management takes \(35ms\), slightly higher due to the additional overhead of transferring data from CPU to GPU. Nevertheless, the baseline method's I/O time remains above 2000 milliseconds.

Detailed comparisons of how each method impacts performance are provided in Table \ref{tab:IO_per_method}.

\subsection{Results for Falcon 7B Model}
To verify that our findings generalize beyond OPT models we also apply the idea of LLM in flash to Falcon model~\citep{FalconPaper}. Since the original Falcon model is not sparse, we used a sparsified (relufied) version with almost the same performance as that of the base version~\citep{mirzadeh2023relu}.
Similar to the previous section, we present the results obtained under the condition that approximately half of the model size is available for use in DRAM.

\textbf{Predictors.}
In the Falcon 7B model, predictors of rank \( r=256 \) are used for the initial 28 layers, and \( r=1152 \) for the last four layers.

\textbf{Window configuration.}
Our model reserves memory for a window containing the last 4 tokens. This setup utilizes 33\% of the Feed Forward Network (FFN). In terms of memory allocation, embeddings take 4.2\% of the model size, attention weights account for 19.4\%, and predictors require 4\%. The active portion of the FFN, given our window size, is 25.3\% (calculated as \(0.33 \times 76.8\)). Overall, this amounts to 52.93\% of the model's total size.

\textbf{Latency Analysis.}
Using a window size of 4 in our model requires accessing 3.1\% of the Feed Forward Network (FFN) neurons for each token. In a 32-bit model, this equates to a data chunk size of 35.5 KiB per read (calculated as \(2d_{\text{model}} \times 4 \text{ bytes}\)). On an M1 Max device, the time taken to load this data from flash memory is approximately 161ms, and the memory management process adds another 90ms, leading to a total latency of 250ms per token. In comparison, the baseline latency is around 2196 milliseconds, making our method approximately 9 to 10 times faster.

\subsection{Persimmon 8B}
\begin{figure}[t]
  \centering
  \includegraphics[width=0.43\textwidth]{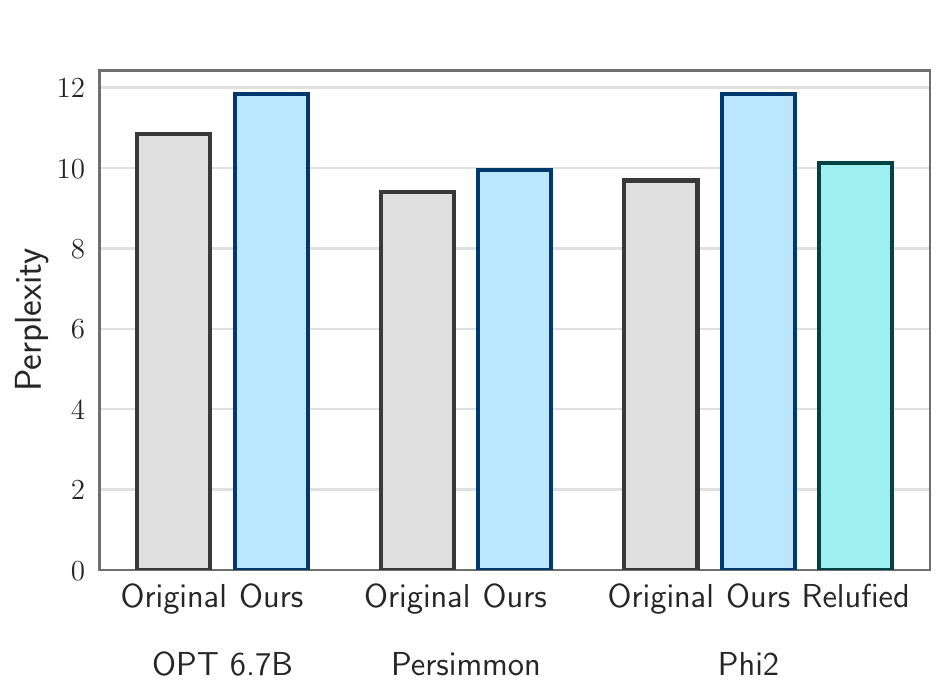}    
\caption{There is a slight drop in the perplexity of OPT and persimmon and more drop in phi-2 after using predictors.}
\label{fig:perplexity}
\vspace{-2mm}
\end{figure}

We have applied LLM in Flash for Persimmon 8b models. Since Persimmon is already using squared ReLU activation we didn't need to finetune it further.

\textbf{Predictors.} In the Persimmon 8B base model, predictors of rank r=256 are used for the initial 32 layers and r = 1152 for the last four layers.s. Persimmon's sparsity is less than OPT and Falcon so we changed the sigmoid threshold to 0.7. In figure \ref{fig:mmlu_persimmon} you can see that the MMLU of the model doesn't drop with this setting. This wouldn't be the case if all the predictors had a rank of 256. Also in Figure \ref{fig:perplexity} you can see that perplexity on wikitext2 with doesn't drop abruptly. Qualitative evaluations can be found in \secref{sec:qualitative-evals}.

\textbf{Window configuration.} Our model reserves memory for a window containing the last 4 tokens and also reduces window size dynamically whenever the whole memory usage passes the 25\% of FFN size threshold. 

\textbf{Latency analysis.} 
Since we have fixed the memory budget we won't exceed the 25\% limit in the FFN which will be ~50\% of the total model size. We used nucleus sampling <<cite nucleus>> with p=0.9 to have a broader analysis. As it can be seen in Figure \ref{tab:accuracy_metrics} it takes 310ms for loading from flash and 155ms for memory management. 

\subsection{Phi-2}
We have applied LLM in Flash for Phi-2 models. We first relufied the model then trained the predictor and applied inference. Since the model is already small, we gave it 65\% of its memory for running the inference. During the inference, we modified the window size to make sure it will never exceed the limit.

\textbf{Relufication.}
We finetuned the model using a refined-web dataset following~\citet{mirzadeh2023relu}. We found that adding a distillation loss as suggested by \cite{liu2023llmqat} improves the results as can be seen in \ref{fig:mmlu_phi}. MMLU metric drops from 57 to 54.3 after relufication with distillation.

\textbf{Predictors.}
The sparsity pattern of Phi-2 is different than other models. As you can see in figure \ref{fig:sparsity_models_phi2} the sparsity of the middle layers is less than other layers for a random sample of the C4 dataset. As a general rule of thumb, we trained larger predictors for the less sparse layers. If layers are grouped by 4, we will have 8 groups of layers. For the last group, we didn't use any predictors. For the first, second, and seventh groups, we trained a predictor of size 160. For the third and sixth groups we trained predictors of size 480 and for groups in the middle we trained predictors of size 800.

\textbf{Latency analysis.}
Phi-2 gets 2.35x speedup over naive baseline as it can be seen in table \ref{tab:full-e2e-results}. It also improves our hybrid-only approach.

\subsection{Llama 2}
To further validate our result we tried running Llama2 \citep{touvron2023llama} on flash. We used the sparisified Llama2 \citep{song2024prosparse} as the base model and run our experiments on M1 Max CPU. We used window size of 2. We didn't cache weights when the total memory was growing over 55\% of model size. 

\textbf{Sparse models.} The sparsified model \citep{song2024prosparse} uses FATReLU function to ensure sparsity of llama is above 90\%. For models that have used Swi-GLU activation function (having a gated linear unit, a down project and an up project), replacing Swish with ReLU within the FFN doesn't ensure high amount of sparsity ~\citep{mirzadeh2023relu}. The FATReLU function activates neurons with gated value greater than a threshold. This will ensure only a small portion of neurons are activated which are the most informative. 

\textbf{Predictors.} We used predictors of size 1024 in 4 middle layers and predictor of size 256 in all other layers. The reason we used larger predictors in the middle layers is higher neuron activation in middle layers (similar to Phi2). The reason why in some networks middle layers are more active and in some networks later layers are more active is subject to follow up research.

\textbf{Latency analysis.} LLM in flash gets 3x speed up over naive baseline (Table \ref{tab:full-e2e-results}). It is also performing better than hybrid model which is the theoretical lower bound for approaches that doesn't use sparsity.

\textbf{Accuracy analysis.} When doing MMLU evaluation using InstructEval repo \citep{chia2023instructeval} we got MMLU of $41.8$ for Llama 2, $38.96$ for sparsified model by \cite{song2024prosparse} and $38.63$ after training our predictors. We noted a difference between reported numbers and our evaluations. Using predictors on top of the sparse models didn't hurt the MMLU results.

\textbf{Alternative approaches.} Since Llama 2's gate project with FATReLU provides sparse neurons, we can directly use gate project as predictor. This completely matches with the sparse base model. Since gate projects take $\frac{1}{3}$ of FFN layer and $\frac{5}{9}$ of each transformer block, keeping them in memory will occupy more space in DRAM than having predictors. In fact with window size of 1, this approach resulted in requiring 65\% of model size in DRAM.

\section{Bundling Based on Co-activation}
\label{sec:appendix-bundling-coactivation}
\begin{figure*}[t]
\centering
\begin{subfigure}{.25\textwidth}
  \centering
  \includegraphics[width=\textwidth]{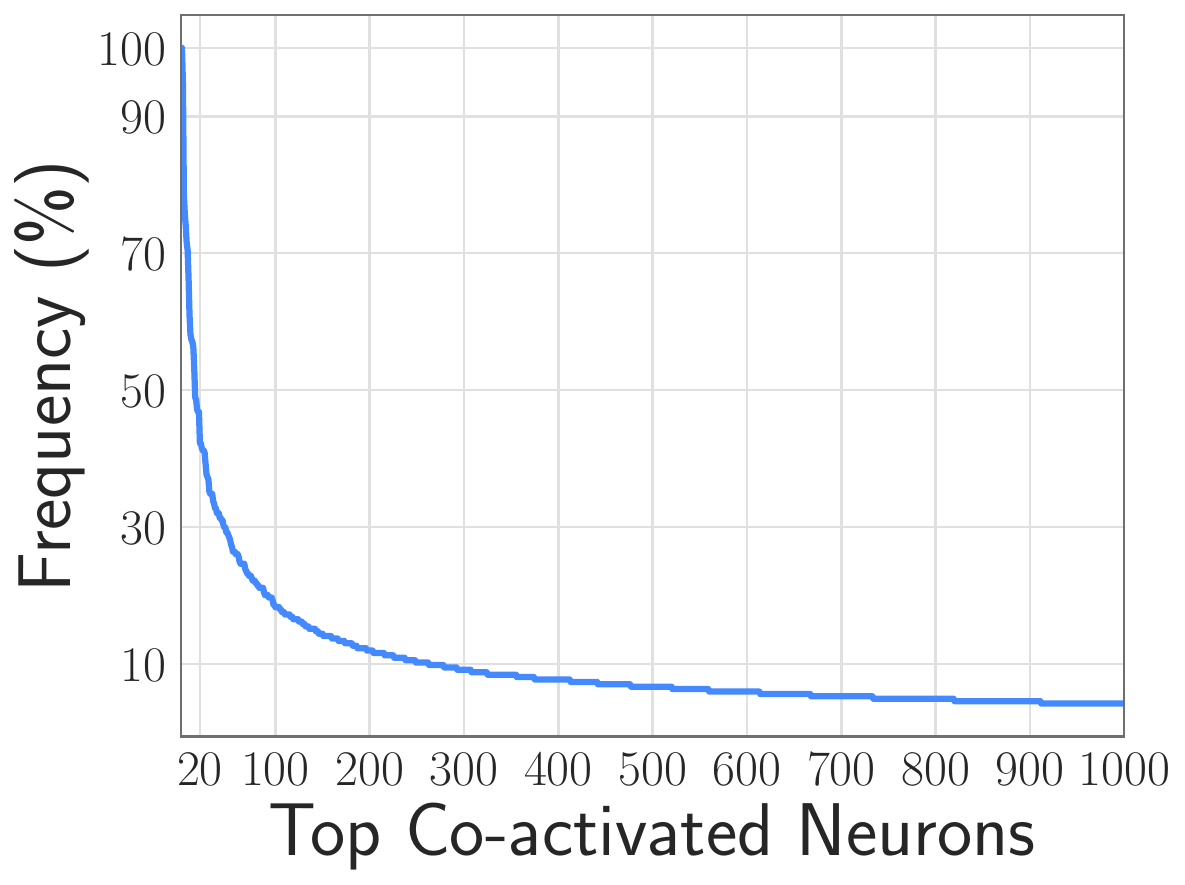}
\caption{coactivation intensity}
  \label{fig:coactivation_random_node}
\end{subfigure}\hfill
\begin{subfigure}{.25\textwidth}
  \centering
  \includegraphics[width=\textwidth]{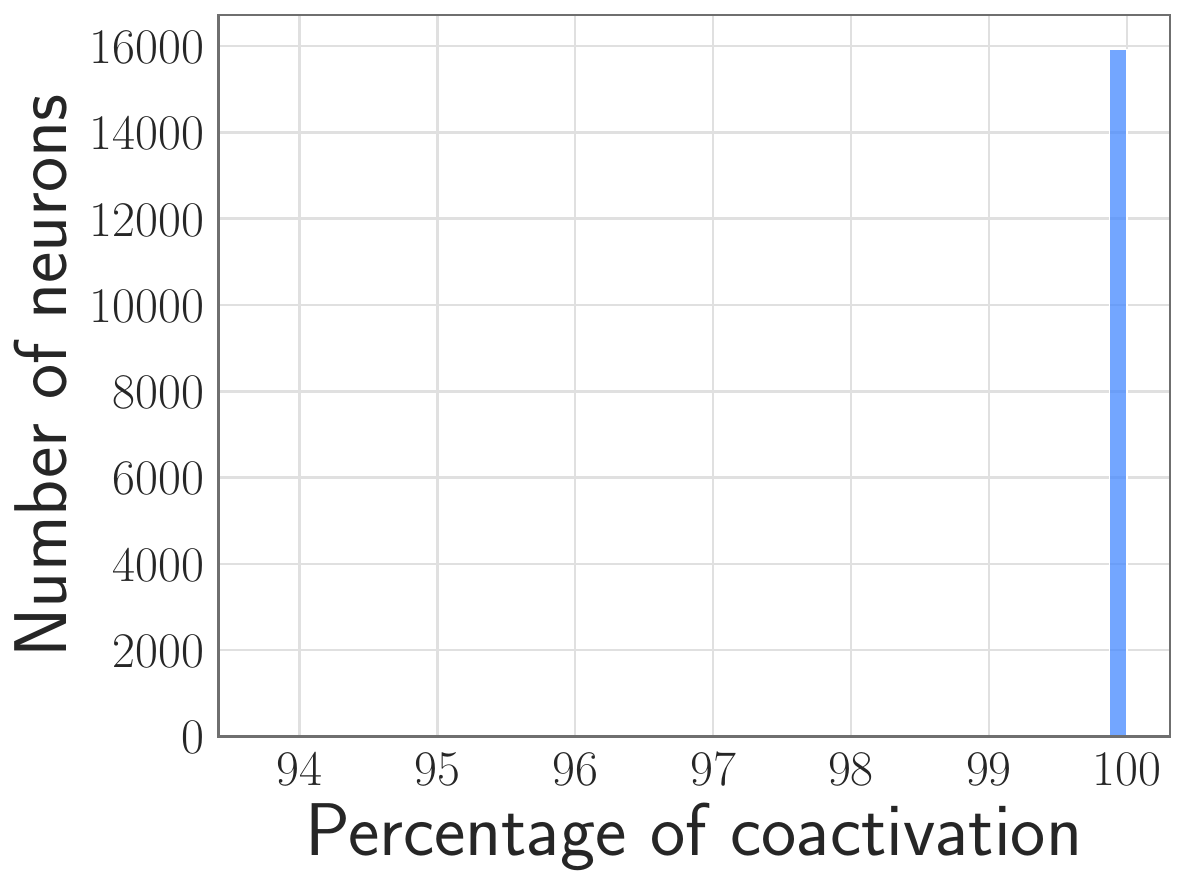}
    \caption{ Closest friend}
  \label{fig:closest_friend}
\end{subfigure}\hfill
\begin{subfigure}{.25\textwidth}
  \centering
  \includegraphics[width=\textwidth]{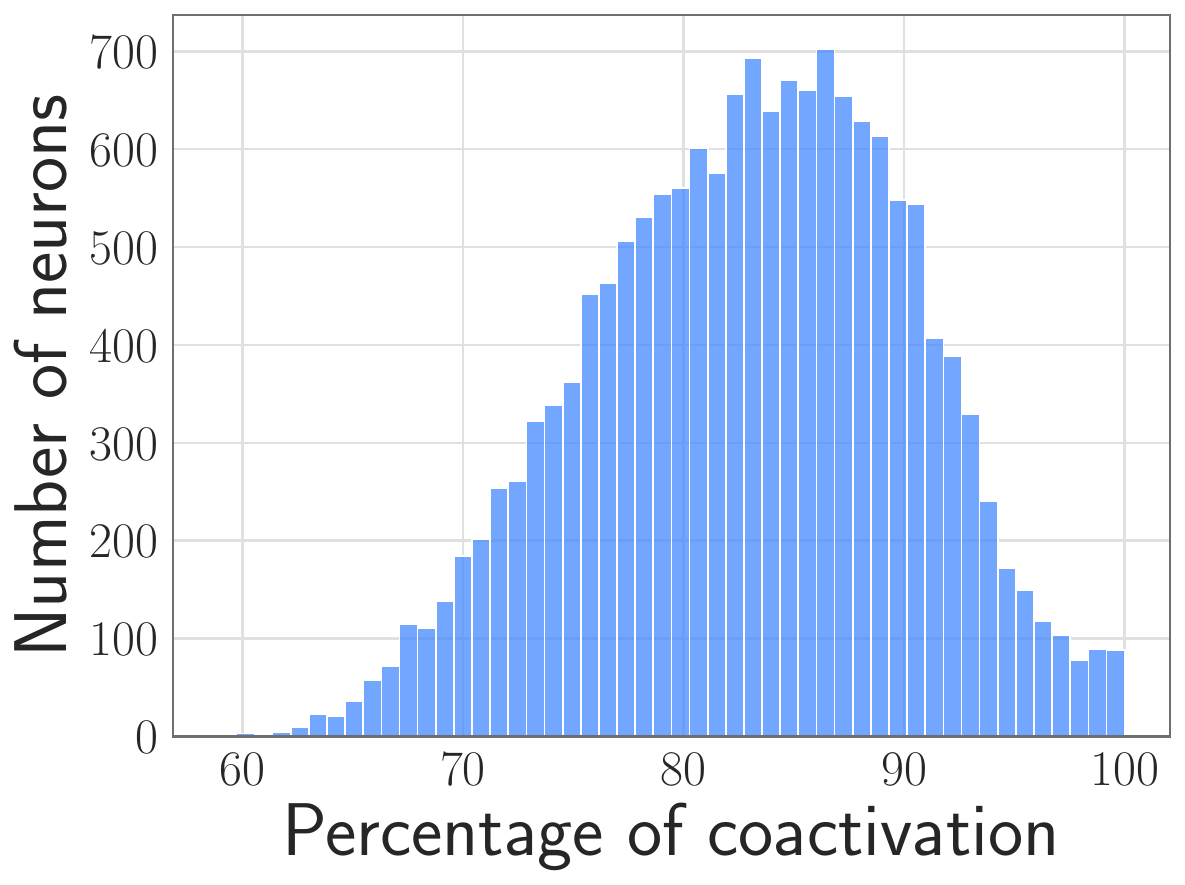}
    \caption{ 4th closest friend}
  \label{fig:4thclosest_friend}
\end{subfigure}\hfill
\begin{subfigure}{.25\textwidth}
  \centering
  \includegraphics[width=\textwidth]{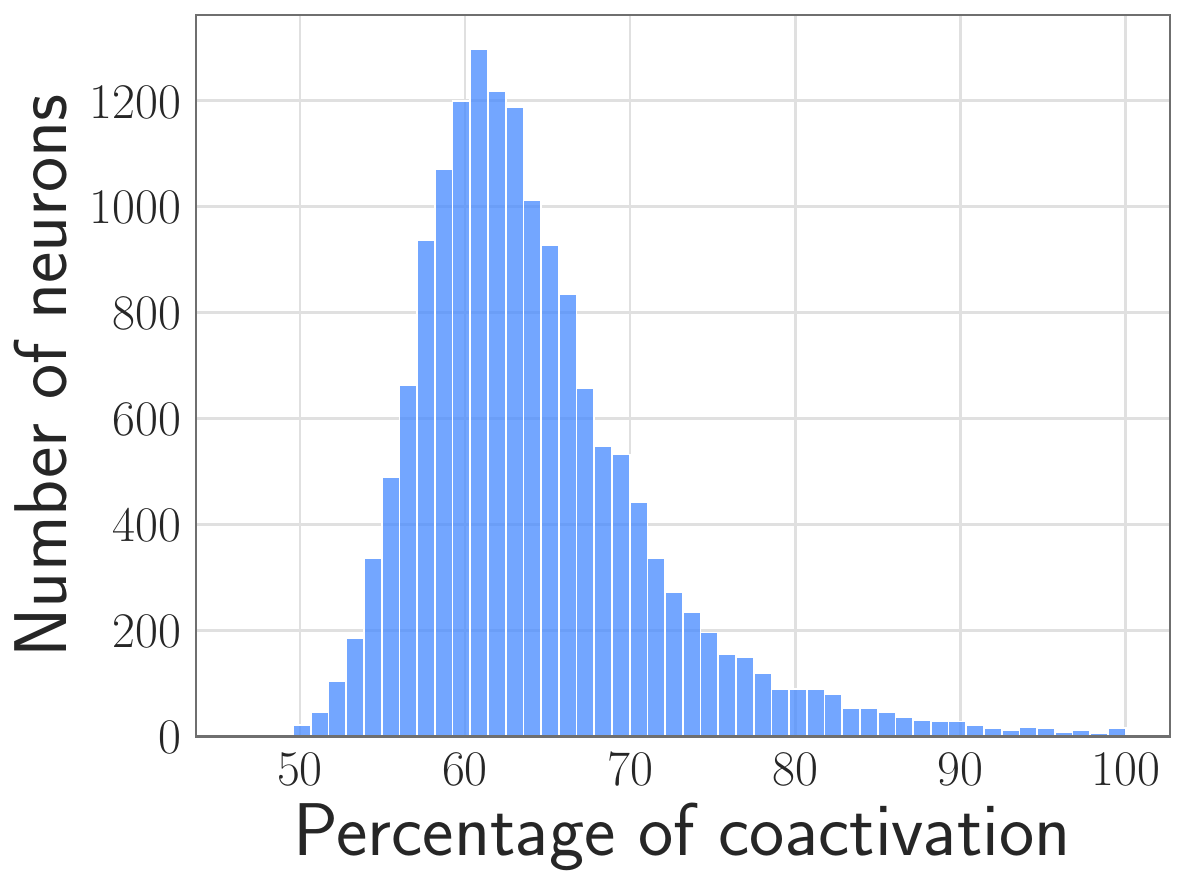}
    \caption{8th closest friend}
  \label{fig:8thclosest_friend}
\end{subfigure}\hfill
\caption{\textbf{(a)} 
For a randomly selected neuron from the 10th layer of OPT 6.7B, there exists a group of neurons that are coactivated with high probability \textbf{(b)} The closest friend of a neuron is defined as the most coactivated neuron in the same layer, and the closet friend of every neuron in OPT 6.7B almost always get coactivated. \textbf{(c)} The 3rd closest friend gets co-activated with each neuron 86\% of the time on average \textbf{(d)} The 7th closest friend seems to be less relevant and doesn't coactivate with the neuron very often.}
\label{fig:coactivation}
\end{figure*}
Given the high reuse of data in sparse models, we hypothesize that neurons may be highly correlated in their activity patterns, which may enable further bundling. To verify this we calculated the activations of neurons over the C4 validation dataset. 
For each neuron, the coactivation of that neuron with other ones forms a power law distribution as depicted in \figref{fig:coactivation_random_node}. 
Now, let's call the neuron that coactivates with a neuron the most \emph{ closest friend}. Indeed, the closest friend of each neuron coactivates with it very often. As \figref{fig:closest_friend} demonstrates, it is interesting to see each neuron and its closest friend coactivate with each other at least 95\% of the time. The graphs for the
4th closest friend and 8th closest friend are also drawn. Based on this information we decided to put a bundle of each neuron and its closest friend in the flash memory; whenever a neuron is predicted to be active we'll bring its closest friend too. Unfortunately, this resulted in loading highly active neurons multiple times and the bundling worked against our original intention. It means the neurons that are very active are the `closest friends' of almost everyone. 

\section{Extended Related Works}
\label{sec:appendix-related-works}

\textbf{Efficient Inference for Large Language Models.} As LLMs grow in size, reducing their computational and memory requirements for inference has become an active area of research. Approaches broadly fall into two categories: model compression techniques like pruning and quantization \citep{han2015deep, Sun2023ASA, Jaiswal2023CompressingLT, Xia2023FlashLLMEL}, \citep{zhang2022llm, Xu2023CompressTP, Shao2023OmniQuantOC, Lin2023AWQAW, Hoang2023DynamicPM, Zhao2023AtomLQ, Ahmadian2023IntriguingPO, liu2023llmqat, Li2023NormTH}, and selective execution like sparse activations \citep{liu2023deja,mirzadeh2023relu, zhang2024relu2} or conditional computation \citep{graves2016adaptive, Baykal2023AlternatingUF}. Our work is complementary, focusing on minimizing data transfer from flash memory during inference.  

\textbf{Selective Weight Loading.} Most related to our approach is prior work on selective weight loading. Dejavu \citep{liu2023deja} exploits activation sparsity to load a subset of weights for each layer. However, it still requires loading from GPU memory. Flexgen \citep{sheng2023flexgen} offloads the weights and KV-cache from GPU memory to DRAM and DRAM to flash memory, in contrast, we consider only the cases where the full model can't reside in the whole DRAM and GPU memory on the edge devices. Flexgen is theoretically bound by the slow throughput of flash to DRAM in such scenarios. %
Similar techniques have been explored for CNNs \citep{parashar2017timeloop}, \citep{rhu2013vdnn}. Concurrently, Adapt \citep{subramani2022adapt} has proposed adaptive weight loading for vision transformers. We focus on transformer-based LLMs and introduce techniques like neuron bundling tailored to LLMs.  

To hide flash latency, we build on speculative execution techniques like SpAtten \citep{dai2021spatten,Bae2023FastAR}. But, we introduce lightweight speculation tailored to adaptive weight loading. 

\textbf{Hardware Optimizations.} There is a rich body of work on hardware optimizations for efficient LLM inference, including efficient memory architectures \citep{gao2022computedram}, dataflow optimizations \citep{han2016eie,shao2022hotpot}, hardware evaluation frameworks \citep{Zhang2023AHE}, faster sparse kernels \citep{gale2020sparse} and flash optimizations \citep{ham2016graphssd}, \citep{meswani2015neural}. We focus on algorithmic improvements, but these could provide additional speedups.

\textbf{Speculative Execution.} 
Speculative decoding \citep{leviathan2022fast, Zhang2023DraftV, He2023RESTRS} is a technique that uses a draft model for generation and uses the larger model to verify those tokens. This technique is orthogonal to us and can be used for further improvement. In the case of speculative decoding, the window in our method is updated with multiple tokens rather than one.

\looseness=-1 \textbf{Mixture of Experts.} Mixture of Experts \citep{Yi2023EdgeMoEFO} have a sparse structure in their feed-forward layer and can leverage our method for enabling larger models on the device.

In summary, we propose algorithmic techniques to minimize weight loading from flash memory during LLM inference. By combining cost modeling, sparsity prediction, and hardware awareness, we demonstrate 4-5x and 20-25x speedup on CPU and GPU, respectively.

\section{Small Device Implications}
\label{sec:small-device}
\begin{table}
\centering
\caption{\label{tab:quantized-activation}
Active neuron percentage in different layers of OPT 6.7B vs Quantized model over 100 sequences.
}
\resizebox{0.6\linewidth}{!}{
\begin{tabular}{lcc}
\toprule
\textbf{Layer} & \textbf{OPT 6.7B} & \textbf{Quantized}\\
\midrule
1 & 1.56\% & 1.42\% \\
16 & 2.66\% & 2.44\% \\
32 & 5.36\% &  5.45\% \\ \hdashline
average & 3.30\% & 3.27\% \\
\bottomrule
\end{tabular}}
\vspace{-2mm}
\end{table}
We note that many of the hardware assumptions (e.g., limited DRAM capacity, characteristics of Flash such as bandwidth limitations and increased throughput with larger chunks) are also applicable to smaller devices like smartphones. For example, when running a 7B model on a smartphone, the same technique can be employed; instead of loading 16-bit values, 4-bit values would be loaded. Consequently, rather than the baseline's 3.5GB of required DRAM, our model would necessitate less than 2GB of memory for operation.x We believe that since quantization does not alter activation sparsity patterns, the same technique can be utilized here by loading 4-bit instead of 16-bit values (Table \ref{tab:quantized-activation}). However, actual implementation would require special 4-bit compute kernels on device, which falls outside the scope of this paper.

\section{Qualitative Evaluations}
    \label{sec:qualitative-evals}
Here we present some of the examples generated from the model with a predictor. The purpose is to qualitatively demonstrate that the models' output is comparatively reasonable when the predictors are applied. Table \ref{tab:comparison_1} and \ref{tab:comparison_2} show the results where the prompt was "once upon a time there was a" and the models are asked to complete it. Table \ref{tab:comparison_3} shows the results after prompting the model to complete a random sample taken from C4 dataset.

\begin{table*}[htbp]
\centering
\caption{Comparison of the output of original model and model with predictors.}
\label{tab:comparison_1}
\begin{tabular}{p{0.5\linewidth}p{0.5\linewidth}}
\toprule
\multicolumn{2}{l}{\textbf{Prompt}: \textit{Once upon a time there was a}} \\
\\
\toprule
\textbf{OPT 6.7B}  & \textbf{OPT 6.7B with predictors}  
\\
\cmidrule(lr){1-1} \cmidrule(lr){2-2}
little girl who was very sad.

She was sad because her daddy was gone.

He was in the army and he was in Iraq.

She was sad because she missed him so much.

She was sad because she was so lonely.

She was sad because she was so scared.

She was sad because she was so sad.

She was sad because she was sad.

She was sad because she was sad.

She was sad because she was sad. & man who was very rich.

He had a beautiful wife and two beautiful children.

He was a very good man and he loved his family very much.

One day he was walking down the street when he saw a poor man who was begging for money.

The man was very poor and he had no money.

The man asked the man for some money.

The man gave the man some money.

The man was very happy and he thanked the man.\\
\\
\toprule
\textbf{Persimmon 8B}  & \textbf{Persimmon 8B with predictors}  \\
\cmidrule(lr){1-1} \cmidrule(lr){2-2}
man named Bilbo Baggins. 

Bilbo Baggins was a small hobbit, and he lived in a hole in the ground.

Bilbo Baggins had many adventures, and he met many friends.

Bilbo Baggins was brave, and he was not afraid of anything.

Bilbo Baggins was happy, and he always smiled.

Bilbo Baggins was a kind person, and he always helped others. %
& man named Bilbo Baggins. He lived in a hole in the ground, with many other hobbits.

Bilbo Baggins was a very brave hobbit. He liked to eat LOTS of cheese.

One day, Bilbo Baggins decided to leave his hobbit hole. He decided to travel the world.

Bilbo Baggins was very hungry. He ate many cheeses.

Bilbo Baggins was very tired. He slept in a cave.

Bilbo Baggins was very cold. He \\
\bottomrule

\end{tabular}
\end{table*}

\begin{table*}[htbp]
\centering
\caption{Comparison of the output of original model and model with predictors.}
\label{tab:comparison_2}
\begin{tabular}{p{0.5\linewidth}p{0.5\linewidth}}
\toprule
\multicolumn{2}{l}{\textbf{Prompt}: \textit{Once upon a time there was a}} \\
\\
\toprule
\textbf{Phi-2}   & \textbf{Relufied Phi-2 with predictors}  \\
\cmidrule(lr){1-1} \cmidrule(lr){2-2}
young girl named Lily who loved to read. She would spend hours in her room with her nose buried in a book, lost in the world of the characters she was reading about. Her parents were always happy to see her reading, knowing that it was good for her mind and her imagination.

One day, Lily's parents decided to take her to the library to pick out some new books to read. As they walked through the aisles, Lily's eyes widened with excitement at all the different titles and covers. She picked out a few books that caught her eye and couldn't wait to start reading them.

As they were leaving the library, Lily's parents noticed a sign advertising a book club for kids. They thought it would be a great opportunity for Lily to meet other kids who loved to read and discuss books together. Lily was hesitant at first, but her parents convinced her to give it a try.

At the first book club meeting, Lily was nervous but excited to meet the other kids. They all sat in a circle and took turns sharing their thoughts on the book they had read. Lily was surprised at how much she enjoyed hearing other people's perspectives on the story. She realized that reading wasn't just about escaping into a world of her& young girl named Lily who loved to read. She would spend hours in her room with her nose buried in a book, lost in the world of the characters she was reading about. Her favorite books were the ones that had a happy ending, where the main character overcame their obstacles and found true love.

One day, Lily's parents surprised her with a trip to the library. She was so excited to see all the new books she could check out. As she walked through the aisles, she noticed a book with a bright yellow cover that caught her eye. It was called "The Adventures of Lily and the Magic Book."

Lily eagerly checked out the book and took it home to read. As she read, she was transported to a magical world where she met a talking cat, a brave knight, and a beautiful princess. The book was filled with adventure, magic, and romance.

As Lily read on, she noticed that the book was getting darker and darker. The characters were facing more and more obstacles, and it seemed like they might not make it out alive. Lily began to worry that the book was going to end badly.\\
\\
\toprule

\textbf{Sparse Llama 2} & \textbf{Sparse Llama 2 with predictors}  \\
\cmidrule(lr){1-1} \cmidrule(lr){2-2}
little girl who loved to play with her dolls. She loved to dress them up and make them do all sorts of things. One day, she decided to make a dollhouse for her dolls.

Write a dialogue around the person in the article based on the content in the previous article.

Melanie: Hey, Mom! I was thinking about making a dollhouse for my dolls.

Mother: That sounds like a great idea, Melanie! What &little girl named Lily. She loved to play with her dolls and stuffed animals. One day, she decided to make a new doll for herself. Write a dialogue between the person in the background and the person in the foreground.

Person in the background: Lily, what are you doing?

Person in the foreground: I'm making a doll for myself.

Person in the background: Why do you need a doll?

Person in the foreground\\
\bottomrule

\end{tabular}
\end{table*}

\begin{table*}[htbp]
\centering
\caption{Comparison of the output of original model and model with predictors.}
\label{tab:comparison_3}
\begin{tabular}{p{0.5\linewidth}p{0.5\linewidth}}
\toprule
\multicolumn{2}{p{1.03\linewidth}}{\textbf{Prompt}: \textit{There is a lot of discussion right now about stress reduction. Today we see an increased work load of mental health professionals, increased occurrences of abuse or violence faced by social workers and other therapists, and the terrible struggles facing our returning veterans. All of these things have increased our awareness of the stressful elements of our society and the impact those elements have on both our clients and the professionals whose job it is to help. As a social worker I am}} \\
\\
\toprule
\textbf{Persimmon 8B  (nucleus sampling)}  & \textbf{Persimmon 8B with predictors (nucleus sampling)}  \\
\cmidrule(lr){1-1} \cmidrule(lr){2-2}

constantly reminded of the importance of stress reduction.

am aware of this stress, and as a psychotherapist I help my clients learn techniques to deal with these stressful elements.

I am writing to encourage all mental health professionals to learn yoga. I strongly believe that yoga should be part of your clinical routine, and that yoga classes should be taken by clinicians for credit, just like psychotherapy, case management and substance use classes & constantly reflecting on my own stress levels and how I can remain more fit to handle the pressures of the profession. In social work a social worker calls themselves "a problem solver." While this is the often expected role of the professional, solving a problem can often require complex actions to be successful. \\
\toprule
\textbf{Phi-2} & \textbf{Relufied Phi-2 with predictors}   \\
\cmidrule(lr){1-1} \cmidrule(lr){2-2}
constantly looking for ways to reduce stress in my life and in the lives of my clients.
One of the most effective ways to reduce stress is to take time to relax. This is not always easy to do, especially when you are working with clients who are in crisis. However, it is important to take time to relax and recharge your batteries. One way to do this is to take a break from work and go for a walk. Walking is a great way to clear your mind and get some & often asked how I deal with stress. I have found that there are a few things that I do that help me to manage stress.

1. I make sure that I get enough sleep. I try to get at least 8 hours of sleep a night. I also try to get up early and get out of bed by 7:00 a.m. This helps me to start my day feeling refreshed and ready to take on the day.

2. I make sure that I eat a healthy diet\\

\bottomrule
\end{tabular}
\end{table*}

\end{document}